\documentclass[runningheads]{llncs}
\usepackage{graphicx}
\usepackage{amsmath,amssymb} 
\usepackage{cleveref}
\usepackage{color}
\usepackage{subcaption}
\begin{document}
\pagestyle{headings}
\mainmatter

\def\ACCV18SubNumber{801}  

\title{Domain Adaptation Meets Disentangled Representation Learning and Style Transfer} 
\titlerunning{Domain Adapt. Meets Disentangled Represent. Learning and Style Transfer}
\authorrunning{Hoang Tran Vu and Ching-Chun Huang}

\author{Hoang Tran Vu and Ching-Chun Huang}
\institute{Department of Electrical Engineering, National Chung Cheng University, Taiwan}

\maketitle

\begin{abstract}
Many methods have been proposed to solve the domain adaptation problem recently. However, the success of them implicitly funds on the assumption that the information of domains are fully transferrable. If the assumption is not satisfied, the effect of negative transfer may degrade domain adaptation. In this paper, a better learning network has been proposed by considering three tasks - domain adaptation, disentangled representation, and style transfer simultaneously. Firstly, the learned features are disentangled into common parts and specific parts. The common parts represent the transferrable features, which are suitable for domain adaptation with less negative transfer. Conversely, the specific parts characterize the unique style of each individual domain. Based on this, the new concept of feature exchange across domains, which can not only enhance the transferability of common features but also be useful for image style transfer, is introduced. These designs allow us to introduce five types of training objectives to realize the three challenging tasks at the same time. The experimental results show that our architecture can be adaptive well to full transfer learning and partial transfer learning upon a well-learned disentangled representation. Besides, the trained network also demonstrates high potential to generate style-transferred images.

\end{abstract}

\section{Introduction}
\label{sect:in}

To decrease the demand on relabeling extra training data when applying the trained network in a new domain, some transfer learning methods for domain adaptation have been proposed in the past decade. Most of the methods aim at reducing the domain shift or minimizing the difference among domain feature distributions. The key idea is to learn deep feature transformations to map inputs from different domains into a common feature space so that the extracted features are both domain-invariant and class-discriminative. Some metrics of domain shift have been used in these methods such as maximum mean discrepancy (MMD) \cite{1,2}, multi kernel variant of MMD (MK-MMD) \cite{3}, and correlation distances \cite{4}. Recently, along with the success and understanding of the Generative Adversarial Network (GAN) \cite{5} for many generative tasks, researchers have turned to borrow the idea of adversarial learning to perform domain adaptation. Based on adversarial learning, domain adaptation problem is modeled as a minimax game between a domain discriminator and a feature extractor. The feature extractor is trained to extract features which can not only minimize the classification loss but also fool the domain discriminator.

Most aforementioned methods focus on learning the common feature representation to bridge and transfer learnable knowledge from a source domain to a target domain \cite{6,7,8,9,10,11}. The success of these methods implicitly funds on the assumption that all information from a source domain are fully transferable. However, some domain specific features are only suitable to characterize the domain properties. Negative effects may happen and degrade domain adaptation if the domain specific features are transferred. Moreover, partial transfer learning \cite{12} is another new concern where the target label space is only a subset of the source label space. Due to the effects of outlier source classes, simplify matching the whole source and target domains as the previous methods would also result in negative transfer. In these cases, learned representation should preferably be disentangled into the transferable and non-transferable features. However, very few studies mention it while solving domain adaptation problems.

\begin{figure}[t]
\centering
\includegraphics[height=5.6cm]{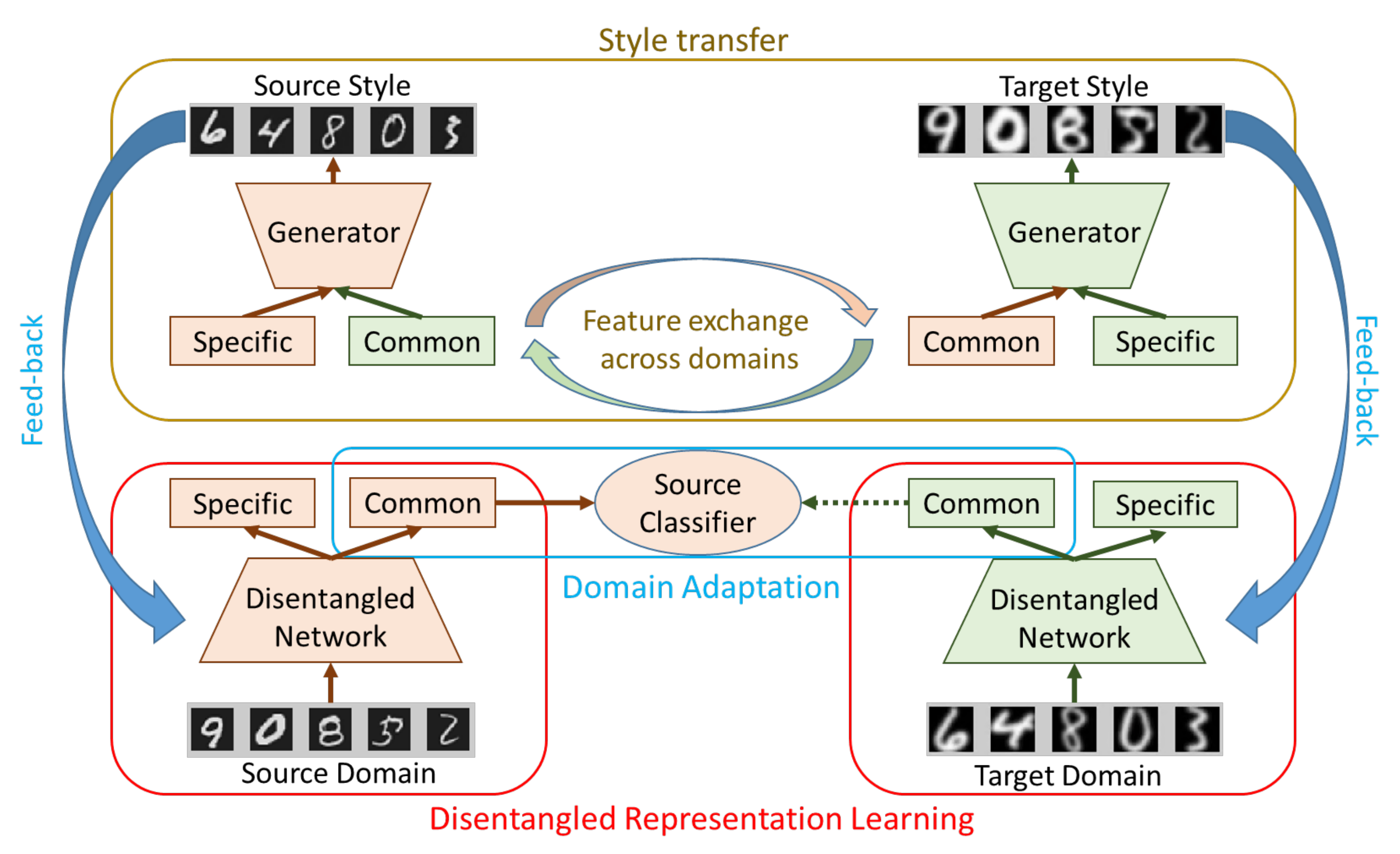}
\caption{Our research objective is to embed three tasks - {\bf Disentangled Representation Learning}, {\bf Domain Adaptation} and {\bf Style Transfer} - into a unified framework, and to discuss their inter-connections. (Bottom) Disentangled Network helps to separate learned features into common parts across domains and domain-specific parts. (Middle) Domain Adaptation is then applied only on the common parts to relieve the negative transfer problem. (Top) Feature exchange across domains allows synthesizing style-transferred images and enhances the transferability of common parts. (Feed-back arrows) the style-transferred images are then feed-back to disentangled network to make sure we can get back the corresponding features. The proposed framework is provided in Fig.~\ref{fig:fig2}.}
\label{fig:intro}
\end{figure}

In this paper, we combine the disentanglement task with domain adaptation, and also introduce the concept of feature exchange across domains that is useful for style-transfer (or image-to-image translation). The designed idea of our framework is illustrated in Fig.~\ref{fig:intro} where three tasks: disentangled representation learning, domain adaptation and style transfer are embbed into a unified framework. Firstly, to relieve the negative transfer problem, we aim to disentangle the domain specific features from the common features which are transferable across the domains. To do so, the designed deep framework divide the representation into two parts: common and specific parts. The common features show the shared contents between two domains. Thus, the domain feature distributions of the common parts are expected to be matched after domain adaption. The specific features mainly reveal the domain properties such as style and appearance. By transferring messages through the common parts rather than the specific parts, we could achieve better transfer learning with less negative propagation. Furthermore, instead of directly ignoring the specific features, we then proposed to integrate style transfer into our domain adaptation framework. As shown in Fig.~\ref{fig:intro}, by exchanging features across domains, the style-transferred images can be synthesized from specific features and exchanged common features. In this paper, we also show how these synthesized images can improve domain adaptation. To sum up, our contributions can be summarized as follows:
\begin{enumerate}
\item {\bf 3-in-one framework}: As far as our knowledge, this is the first work discussing about the relationship of three tasks: disentangled representation learning, domain adaptation and style transfer and combining them into a unified framework.
\item {\bf Disentangled Representation Learning for Domain Adaptation}: By disentangling the features, we aim to decompose the feature space into common parts which are suitable for domain transfer and specific parts which are not transferable. Therefore, we can reduce the influence of negative transfer.
\item {\bf Feature Exchange for Style Transfer}: We propose a new concept of feature/style exchange across domains. In our framework, the specific part of one domain can be combined with the common part of the another one to form a new feature representation that later on can be used to generate a style-transferred image through a GAN-based framework. With this idea, our framework can not only enhance the transferability of common features but also be extended to solve multi-domain image-to-image translation problems.
\item {\bf Feed-back design}: Based on the ability to generate a style-transferred image, we propose a novel feed-back design and a semantic consistency loss to enhance the transferability of the learned common parts and preserve the semantic information during feature exchange process.
\end{enumerate}

In brief, the paper organization is as follows: Section~\ref{sect:rw} summarizes the related works. Section 3 presents the proposed method. Section 4 presents the datasets and the experimental results. Finally, we conclude this work in Section 5.

\section{Related Works}
\label{sect:rw}
{\bf Domain Adaptation}: A large number of domain adaptation methods have been proposed over the recent years. We divide them into two main groups: 1) finding the mapping between source and target domains, and 2) finding the shared latent space between two domains. For the first group, some methods focus on learning the mapping from a source domain to a target domain in the feature level \cite{7} or the image level \cite{8}. Some other methods choose to learn the inverse mapping from target to source \cite{6}. Recently, to make the adaptation system more robust and general, the authors in \cite{9} proposed to combine both mapping directions in a unified architecture. For the second group, researchers focus on minimizing the distance between two domains in feature space by using first order statistic \cite{1,2}, or higher order statistic \cite{3,4}. More recently, domain shift between two domains could be further reduced by applying adversarial learning. The main concept is to find the shared latent space so that a strong domain classifier could not distinguish source samples and target samples. There are variants of training strategies for adversarial learning. In \cite{10}, the authors proposed to combine a domain confusion loss and softmax cross-entropy loss to train the network so that the network could transfer knowledge not only across domains but also tasks. However, the method still needs sparsely labeled data in target domain. In \cite{6}, the authors proposed an unsupervised adversarial discriminative domain adaptation framework. According to their result, an adversarial learning framework based on asymmetric feature mappings for source and target can outperform the one based on a symmetric mapping. To realize adversarial learning, the choice of adversarial loss function is another issue. In \cite{11}, the authors directly treat the domain discriminator loss as the adversarial loss to learn the optimal feature mapping; while the authors in \cite{6} train the optimal feature mapping with the standard discriminator loss function with inverted labels.
 
{\bf Disentangled Representation Learning}: Recently, the research group in \cite{13} argued that most of the conventional domain adaptation methods learn the common representations of source and target domains without considering the negative influence from the domain specific characteristics. If the network transfers the negative effects, we may not be able to learn a well generalized common feature representation. Therefore, the authors proposed a two-stage neural network learning algorithm to learn a multi-part hidden layer where individual parts can be disentangled or combined for different tasks in different domains. Also aiming at answering the fundamental question “{\bf{\emph{what to transfer}}}”, Domain Separation Network \cite{14}, proposed to integrate a private network to learn the private subspace for each domain and a shared network to extract the shared representation subspace across domains. To force the private subspace and shared subspace to be independent, a difference loss is introduced. After domain separation, the standard adversarial loss and classification loss can be applied only in the shared subspace.

{\bf Partial Transfer Learning}: Partial transfer learning was proposed in \cite{12}, where the target domain label space is a subspace of the source domain label space. Because the extra source classes might cause negative transfer when classifying the target domain, it makes the domain adaptation problem more challenging. In this work, in order to solve the partial transfer problem, instead of using single-discriminator domain adversarial network, the authors proposed to use multi-discriminator domain adversarial network, each discriminator is responsible for matching the source and target domain data associated with each label.

\begin{figure}[t]
\centering
\includegraphics[height=5cm]{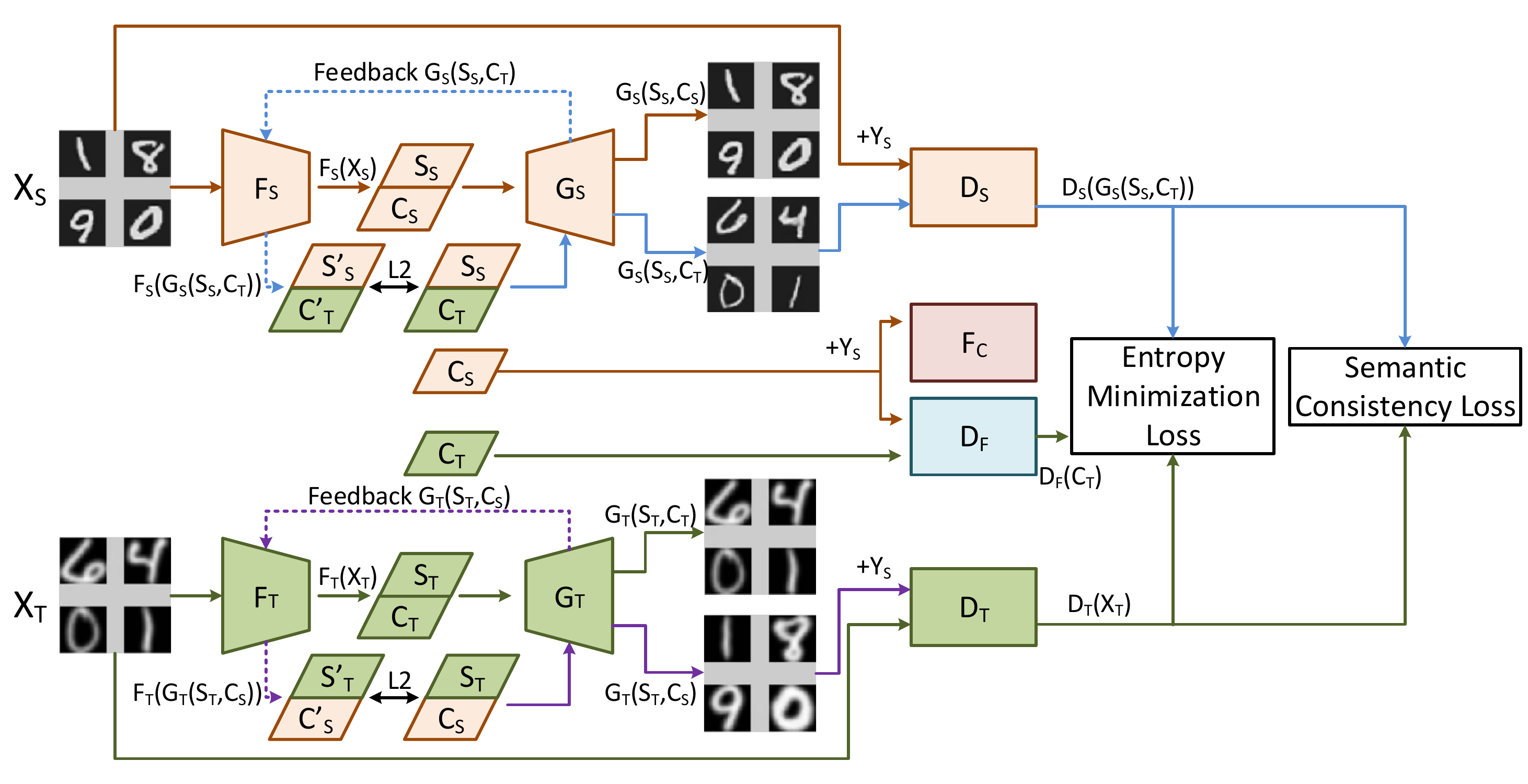}
\caption{Proposed framework.}
\label{fig:fig2}
\end{figure}

\section{The Proposed Domain Adaptation Network}

In this paper, we firstly focus on the problem of unsupervised domain adaptation. Particularly, we aim to train a target domain classifier $F_C$ that can correctly predict the label for the target data {\it $X_T$} by leveraging the source data {\it $X_S$} and source labels {\it $Y_S$}. Unlike the conventional setting of domain adaptation, we pay further attention on (a) transfer learning of common feature components and (b) partial transfer learning. For case (a), we argue that disentangled representation learning is necessary to decompose the common feature parts and specific parts of domains. For case (b), we design our framework to match the cluster-based distributions between source and target domains instead of matching the whole domain distribution. Finally, we proposed to exchange the specific feature parts across domains and synthesize style-transferred images. These style-transferred images are further used to improve domain adaptation. Our framework is illustrated in Fig.~\ref{fig:fig2}. Two key function modules in our feature extraction and image generation framework are summarized as follows:
\begin{itemize}
\item {\bf Disentangled Representation Learning}: The feature extraction networks, {\it $F_S$} and {\it $F_T$}, have the ability to disentangle the feature maps into different semantic parts including the common parts ({\it $C_S$} and {\it $C_T$}) and specific parts ({\it $S_S$} and {\it $S_T$}). We denote the decomposition processes as {\it $C_S=F_S^C(x_S), C_T=F_T^C(x_T), S_S=F_S^S(x_S)$}, and {\it $S_T=F_T^S(x_T)$}. As an instance, in our digit classification experiments, the common parts imply the semantic features of “digits”; the specific parts mainly reveal the domain style such as the writing style and appearance. Each part could be utilized in a different way to connect two domains and finally helps to transfer knowledge for domain adaptation.
\item {\bf Feature exchange across domains}: We combine the specific part of one domain with the common part of the other one to synthesize a new feature representation as shown in Fig.~\ref{fig:fig2}. Next, based on a learnable GAN-based network, our system can generate a style-transferred image given a synthesized feature representation. The style-transferred image, keeping the original image content but different image style, would play important roles to bridge domains. This process also enable our system to transfer the labels from source to target.
\end{itemize}

Besides feature extractors and generators, our framework also consists of three discriminators and one standard classifier as illustrated in Figure~\ref{fig:fig2}. Based on the supervised labels in the source domain, the standard classifier ( {\it $F_C$}) and the source common features should be well trained in order to correctly classify the source domain samples. In addition, three adversarial discriminators  {\it $D_S$, $D_T$} and  {\it $D_F$} are introduced for representation adaptation in both the image level and feature level.  {\it $D_S$} aims to distinguish between real source images {\it $X_S$} and style-transferred target image, denoted as {\it $\left\{G_S(S_S,C_T)\right\}$}. In the similar way, {\it $D_T$} aims to discriminate between {\it $X_T$} and {\it $\left\{G_T(S_T,C_S)\right\}$}. By applying adversarial learning in the image level with {\it $D_S$} and {\it $D_T$}, we hope that the style-transferred images should not be distinguished from the real ones. This property is quite helpful for domain adaptation, which would be explained later. Moreover, we also apply adversarial learning in the feature level by {\it $D_F$}, which is designed to distinguish between the common parts {\it $C_S$} and {\it $C_T$} of source and target domains. With the helps of the adversarial discriminators {\it $D_F$}, we can match the feature distribution of the source common part {\it $C_S$} and the target common part {\it $C_T$}. To train our domain adaptation network, we designed and implemented 5 types of objective losses as described in the following subsections.

\subsection{Adversarial Losses}

Our domain adaptation framework works like the conditional GANs \cite{17} in which the conditional variables are common parts {\it C} and specific parts {\it S}. However, we do not use any random noise as GAN inputs. All the features come from data themselves. For domain adaptation, we apply adversarial losses \cite{15} and design three adversarial discriminators ({\it $D_S, D_T, D_F$}) to jointly match the distributions of two domains in both the image and feature levels. However, our discriminators not only aim to distinguish between source and target domains but also try to classify the input data. Similar to the design in \cite{18}, our adversarial discriminators are ({\it $N_C+1$})-way classifiers with {\it $N_C$} binary nodes to indicate {\it $N_C$} content classes and an extra class for type discrimination. For {\it $D_S$} and {\it $D_T$}, the extra class represents a real (1) or synthesized (0) image; For {\it $D_F$}, it indicates the source (1) or target (0) domain. With this design, the discriminators could be more powerful and help our system to match the cluster-based distribution between domains instead of matching the whole domain distribution as the traditional discriminator \cite{5}. Note that, due to image style transfer, our network can still train {\it $D_T$} even though we do not have supervised labels in the target domain. Indeed, the label information is borrowed from the annotation in the source domain.
Accordingly, we have defined three adversarial losses for training: (a) the feature level loss {\it $L_{(adv\_fea)}$}, (b) the image-level loss in the source domain {\it $L_{(adv\_img)}^S$}, and (c) the image-level loss in the target domain {\it $L_{(adv\_img)}^T$}.

\subsubsection{Feature Level Loss ({\it $L_{adv\_fea}$})}
Since our discriminator plays two roles at the same time, a type classifier and a content classifier, {\it $L_{adv\_fea}$} is composed of a type loss {\it $L_{type\_D_F}$} and a classification loss {\it $L_{cls\_D_F}$} as defined in equation~\ref{eq:eq1}.
\begin{equation} 
\label{eq:eq1}
L_{adv\_fea}(F_S, F_T, D_F)= L_{type\_D_F}(F_S, F_T, D_F) + L_{cls\_D_F}(F_S, D_F).
\end{equation}
In (\ref{eq:eq1}), the type loss {\it $L_{type\_D_F}$} is optimized by the following function:
\begin{equation} 
\label{eq:eq2}
\min_{F_S, F_T}\max_{D_F} E_{x_S}log(D_F^{N_C+1}(F_S^C(x_S))) + E_{x_T} log(1-D_F^{N_C+1}(F_T^C (x_T))).
\end{equation}
Here, {\it $D_F^{(N_C+1)}$}(.), the {\it $(N_C+1)^{th}$} output of discriminator {\it $D_F$}, works as a domain classifier and predicts the domain label (1 for source and 0 for target) for each input sample. {\it $F_S^C (x_S )$} and {\it $F_T^C (x_T)$} are the common feature parts of a source sample {\it $x_S$} and a target sample {\it $x_T$}. On the other hand, the classification loss {\it $L_{cls\_D_F}$} is defined as a standard softmax cross-entropy. That is,
\begin{equation} 
\label{eq:eq3}
L_{cls\_D_F}(F_S, D_F)=\min_{F_S, D_F} -E_{x_S}\sum_{i=0}^{N_C}y_S^i.log(\sigma(D_F^i(F_S^C(x_S)))).
\end{equation}
where {\it $D_F^i$}(.), the {\it $i^{th}$} output of discriminator {\it $D_F$}, works as a content classifier and predicts the probability of the {\it $i^{th}$} class. {\it $y_S^i$} is the supervised label for the input sample {\it $x_S$}. In (\ref{eq:eq3}), {\it $\sigma$}(.) denotes the softmax function.

\subsubsection{Source Domain Image-level Loss ({\it $L_{adv\_img}^S$})}
Like {\it $L_{adv\_fea}$}, {\it $L_{adv\_img}^S$} is also composed of a type loss {\it $L_{type\_D_S}$} and a classification loss {\it $L_{cls\_D_S}$}. That is
\begin{equation} 
\label{eq:eq4}
L_{adv\_img}^S(F_S, F_T, G_S, D_S)= L_{type\_D_S}(F_S, F_T, G_S, D_S) + L_{cls\_D_S}(D_S),
\end{equation}

The type loss {\it $L_{type\_D_S}$} related to discriminator {\it $D_S$} is also defined by traditional adversarial loss function. That is
\begin{equation}
\label{eq:eq5}
\min_{F_S, F_T, G_S}\max_{D_S}E_{x_S}log(D_S^{N_C+1}(x_S)) + E_{S_S,C_T}log(1-D_S^{N_C+1}(G_S(S_S,C_T))).
\end{equation}
In (\ref{eq:eq5}), {\it $D_S^{(N_C+1)}$}(.) is a type classifier used to discriminate a real or synthesized image. {\it $G_S (S_S, C_T)$} generates an image by concatenating a source specific feature {\it $S_S$} and a target common feature {\it $C_T$}. Note that {\it $S_S=F_S^S(x_S)$} and {\it $C_T=F_T^C (x_T)$}.

Similar to (\ref{eq:eq3}), {\it $L_{cls\_D_S}(D_S)$}, is defined as in equation (\ref{eq:eq6}).
\begin{equation} 
\label{eq:eq6}
L_{cls\_D_S}(D_S)=\min_{D_S} -E_{x_S}\sum_{i=0}^{N_C}y_S^i.log(\sigma(D_S^i(x_S))).
\end{equation}

\subsubsection{Target Domain Image-level Loss ({\it $L_{adv\_img}^T$})}
Target adversarial loss {\it $L_{adv\_img}^T$} shares the same design concept as {\it $L_{adv\_img}^S$}. It is defined as
\begin{equation} 
\label{eq:eq7}
L_{adv\_img}^T(F_S, F_T, G_T, D_T)= L_{type\_D_T}(F_S, F_T, G_T, D_T) + L_{cls\_D_T}(F_S, F_T, G_T, D_T),
\end{equation}
where the type loss {\it $L_{type\_D_T}$} and image content classification loss {\it $L_{cls\_D_T}$} are
\begin{multline}
\label{eq:eq8}
L_{type\_D_T}(F_S, F_T, G_T, D_T)=\\
E_{x_T}log(D_T^{N_C+1}(x_T)) + E_{S_T,C_S}log(1-D_T^{N_C+1}(G_T(S_T,C_S))).
\end{multline}
\begin{equation} 
\label{eq:eq9}
L_{cls\_D_T}(F_S, F_T, G_T, D_T)= -E_{S_T,C_S}\sum_{i=0}^{N_C}y_S^i.log(\sigma(D_T^i(G_T(S_T,C_S)))).
\end{equation}
Here, it is worth mentioning that the supervised class label {\it $y_S^i$} for training the target discriminator {\it $D_T$} is borrowed from the source domain. In (\ref{eq:eq9}), we generate many target-style images with source labels so that {\it $D_T$} could be trained.

\subsection{Feedback Losses and Reconstruction Losses}

Because our framework is designed in the form of a convolutional auto-encoder, in order to make sure the learned features are generalized enough, we also proposed to minimize the reconstruction losses for both domains as shown in (\ref{eq:eq10}).
\begin{equation} 
\label{eq:eq10}
\min_{G_S,F_S,G_T,F_T}E_{x_S}||x_S-G_S(F_S(x_S))||_2^2+E_{x_T}||x_T-G_T(F_T(x_T))||_2^2.
\end{equation}

Besides, inspired by the cycle-consistency loss \cite{19}, we proposed feedback losses to enforce the learning of feature extractors and image generators. In detail, we input a combined feature map ({\it $S_S, C_T$}) into a generator {\it $G_S$} to generate a source-style image. Ideally, if we input the synthesized image into feature extractor {\it $F_S$} and get its feature map {\it $S_S', C_T'$}, we would hope the two feature maps {\it $S_S', C_T'$} and {\it $S_S, C_T$} are consistent. The similar concept could also be applied to the target domain. These {\bf \emph {feedback}} constraints could be integrated in our learning step by imposing an L2 penalty term according to the feedback errors as in (\ref{eq:eq12}) and (\ref{eq:eq13}). 
\begin{multline} 
\label{eq:eq12}
L_{feedback}^S(F_S, F_T, G_S)= \\
E_{S_S, C_T}(||S_S-F_S^S(G_S(S_S,C_T))||_2^2+||C_T-F_S^C(G_S(S_S,C_T))||_2^2).
\end{multline}
\begin{multline} 
\label{eq:eq13}
L_{feedback}^T(F_S, F_T, G_T)= \\
E_{S_T, C_S}(||S_T-F_T^S(G_T(S_T,C_S))||_2^2+||C_S-F_T^C(G_T(S_T,C_S))||_2^2).
\end{multline}

\subsection {Semantic consistency loss}
Our framework allows the generated target images to inherit the source labels and enables the transfer of sample labels. Furthermore, based on the supervised label transfer, discriminators {\it $D_S$} and {\it $D_T$} could be trained. By leveraging {\it $D_S$} and {\it $D_T$}, we proposed a new semantic consistency loss to improve domain adaptation. If our network is well trained, we expect the classification result {\it $D_S(G_S(S_S, C_T|_{X_T})$} of the generated style-transferred image {\it $G_S(S_S, C_T|_{X_T})$} should be consistent with {\it $D_T(X_T)$}, where {\it $C_T|_{X_T}$} is the common feature vector extracted from sample {\it $X_T$}. Therefore, in order to encourage this kind of semantic consistency, we introduce the {\bf \emph{}semantic consistency loss} as follows:
\begin{equation} 
\label{eq:eq14}
L_{Sem}(F_S, F_T, G_S, D_S, D_T)= E_{S_S,X_T}||D_T^{1\rightarrow N_C}(X_T)-D_S^{1\rightarrow N_C}(G_S(S_S, C_T|_{X_T}))||_2^2.
\end{equation}

{\bf \emph {Semantic consistency loss}} plays the important role to connect both domains. To reduce the {\bf \emph {semantic consistency loss}}, our feature extraction networks {\it F} are forced to well disentangle the Common and Specific feature components. Meanwhile, the generators {\it G} are forced to perform style transfer well.

\subsection {Entropy minimization losses}
The classification ability of the discriminators {\it $D_S, D_T$} and {\it $D_F$} is a critical point in our network. However, so far, we only base on the source sample annotation for training. The result might be acceptable for {\it $D_S$} and {\it $D_F$} but might not be perfect for {\it $D_T$} due to the lack of true labels. Thus, to enhance the classification ability, we look for the help from unsupervised methods \cite{20} and integrate the concept of entropy minimization \cite{16} into our network training.

If we treat the output vector of {\it $D_*^{1\rightarrow N_C}\in \left\{D_S^{1\rightarrow N_C}, D_T^{1\rightarrow N_C}, D_F^{1\rightarrow N_C}\right\}$} as a probability distribution, its entropy can be measured by
\begin{equation} 
\label{eq:eq15}
H(\sigma(D_*^{1\rightarrow N_C}(x_*^i)))= -\sum_{j=1}^{N_C}\sigma(D_*^j(x_*^i)).log(\sigma(D_*^j(x_*^i))),
\end{equation}
where {\it $N_C$} is the class number and {\it $\sigma$}(.) denotes the softmax function. For an input sample {\it $x_*^i$}, if the corresponding entropy is small, it implicitly means the sample is well classified from an unsupervised viewpoint. Thus, we might enhance the classification ability by minimize the summarization of the entropies of many samples. To utilize this property in our training, we define the three entropy loss terms corresponding to three discriminators {\it $D_S, D_T$}, and {\it $D_F$} as follows:
\begin{multline} 
\label{eq:eq16}
L_{Entropy}(F_S, F_T, G_S, D_S, D_T, D_F)= E_{S_S, C_T}H(\sigma(D_S^{1\rightarrow N_C}(G_S(s_S^i,c_T^i))) \\
+E_{x_T}H(\sigma(D_T^{1\rightarrow N_C}(x_T^i))+E_{C_T}H(\sigma(D_F^{1\rightarrow N_C}(c_T^i)).
\end{multline}

By minimizing the entropy penalty in (\ref{eq:eq16}), our system has two achievements. (1) We can train the three discriminators, the classification part, by unlabeled target samples. (2) The feature extractors are trained to from a well-clustered feature distribution over many classes. These properties improve domain adaptation especially in the partial transfer case. To understand the importance of entropy minimization, we visualize the learned features when training with and without {\it $L_{Entropy}$} in Fig.~\ref{fig:fig5b} and Fig.~\ref{fig:fig5c}. We may find the margins among the clusters are clear and well separated when the effect of {\it $L_{Entropy}$} is considered in the training process.

\subsection {Classification loss}
The last loss we apply in our learning is the standard {\it classification loss}. It uses the labeled source samples to train the classifier {\it $F_C$} in the common feature domain and predict the final output label for a given testing sample. For the {\it $N_C$}-way classification, the multiple-class classification loss are defined as
\begin{equation} 
\label{eq:eq17}
L_{cls}(F_S, F_C)=\min_{F_S, F_C} -E_{C_S}\sum_{i=0}^{N_C}y_S^i.log(\sigma(F_C^i(C_S))).
\end{equation}

\section{Experiments and Discussions}
\label{sec:exp}

\subsection{Setup}
In order to evaluate the effectiveness of our framework, we validate it by performing domain adaptation on three standard digit datasets: MNIST \cite{21}, USPS \cite{22}, and SVHN \cite{23} which contain 10 classes of digits; and two traffic sign datasets: Syn-Signs \cite{11} and GTSRB \cite{31} which contain 43 classes of traffic signs. The details of these datasets are described in Supplemental Material A.

With these datasets, we take into account the following unsupervised transfer scenarios. (1) {\bf MNIST $\rightarrow$ USPS}:  because {\bf USPS} and {\bf MNIST} follow very different distribution, to accelerate experiments, we follow the training protocol created in \cite{24}, randomly sampling 1800 images in {\bf USPS} and 2000 images in {\bf MNIST}. To reduce the high variance effect in performance of random sampling, we run each experiment five times and report the average performance. (2) {\bf SVHN $\rightarrow$ MNIST}: we use the full training sets. All images were rescaled into 32x32 and pixels were normalized to [0, 1] values, and only the labels from source are available during training. (3) {\bf Syn-Signs $\rightarrow$ GTSRB}: we also use full training sets. The region of interest around the sign is extracted and rescaled into 40x40 from each image in GTSRB to match those of Syn-Signs.

Two transfer learning problems are focused in our experiments are full transfer learning and partial transfer learning where the target label space is a subset of source label space. For the case of partial transfer learning, we randomly select 5 classes to form the target domain data, and only the scenario {\bf MNIST $\rightarrow$ USPS} is considered.

{\bf Architecture}. For all of these experiments, we simply modify LeNet architecture provided in Caffe source code \cite{25} as our extractors, and use the same structure as DCGAN \cite{26} for the generators. The details of network architectures are given in Supplemental Material B, the training procedures and hyperparameters are discussed in Supplemental Material C.

\setlength{\tabcolsep}{4pt}
\begin{table}[t]
\begin{center}
\caption{Experimental results on unsupervised adaptation}
\label{table:tb1}
\begin{tabular}{p{2.5 cm} | p{2cm} | p{1.8cm} | p{2.3cm}}
\hline\noalign{\smallskip}
 						& {\bf MNIST to USPS} 			& {\bf SVHN to MNIST}	& {\bf Syn-Signs to GTSRB}	\\
\noalign{\smallskip}
\hline
\noalign{\smallskip}
Source only 			& 78.9 								& 60.1 \underline+ 1.1 	& 79.0							\\
\hline
CORAL \cite{4} 		& 81.7 								& 63.1 						& 86.9							\\
MMD \cite{2} 			& 81.1 								& 71.1						& 91.1							\\
DANN \cite{7} 			& 85.1								& 73.9						& 88.7							\\
DSN \cite{14} 			& 91.3 								& 82.7 						& 93.1							\\
CoGAN \cite{28} 		& 91.2 								& No converge				& 								\\
ADDA \cite{6}			& 89.4 \underline+ 0.2 			& 76.0 + 1.8				&								\\
GenToAdapt \cite{18} 	& 92.5 \underline+ 0.7 			& 84.7 \underline+ 0.9 	&								\\
DRCN \cite{29} 		& 91.8 \underline+ 0.09 			& 82.0 \underline+ 0.16 	&								\\
\hline
{\bf Our method} 		& {\bf 94.14} 						& {\bf 90.23}				& {\bf 94.66}					\\
\hline
\end{tabular}
\end{center}
\end{table}

\begin{figure}[h]
\begin{center}
\begin{subfigure}{0.45\textwidth}
\includegraphics[width=0.9\linewidth, height=4.5cm]{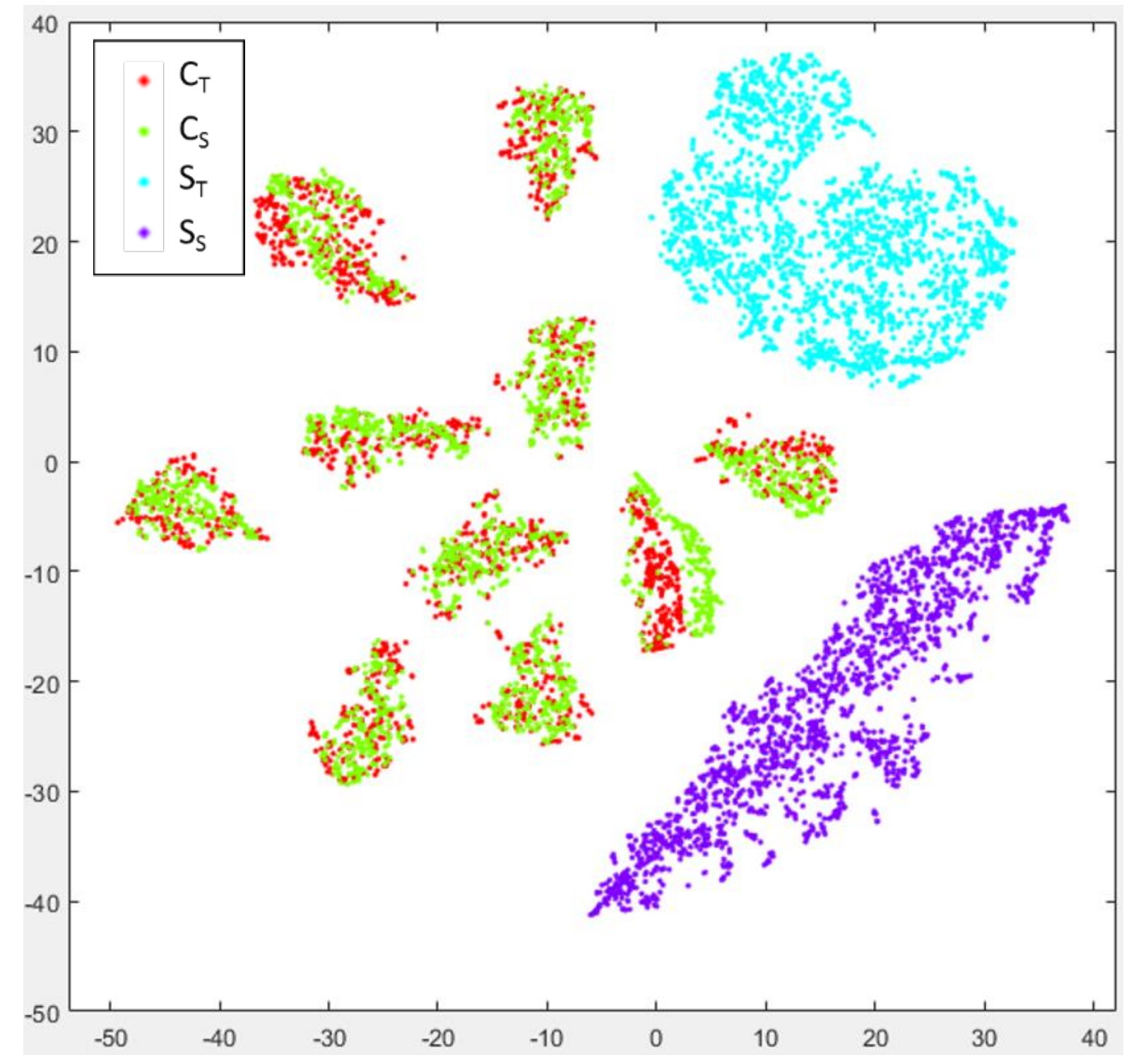} 
\caption{{\bf MNIST $\rightarrow$ USPS}}
\label{fig:fig3b}
\end{subfigure}
\begin{subfigure}{0.45\textwidth}
\includegraphics[width=0.9\linewidth, height=4.5cm]{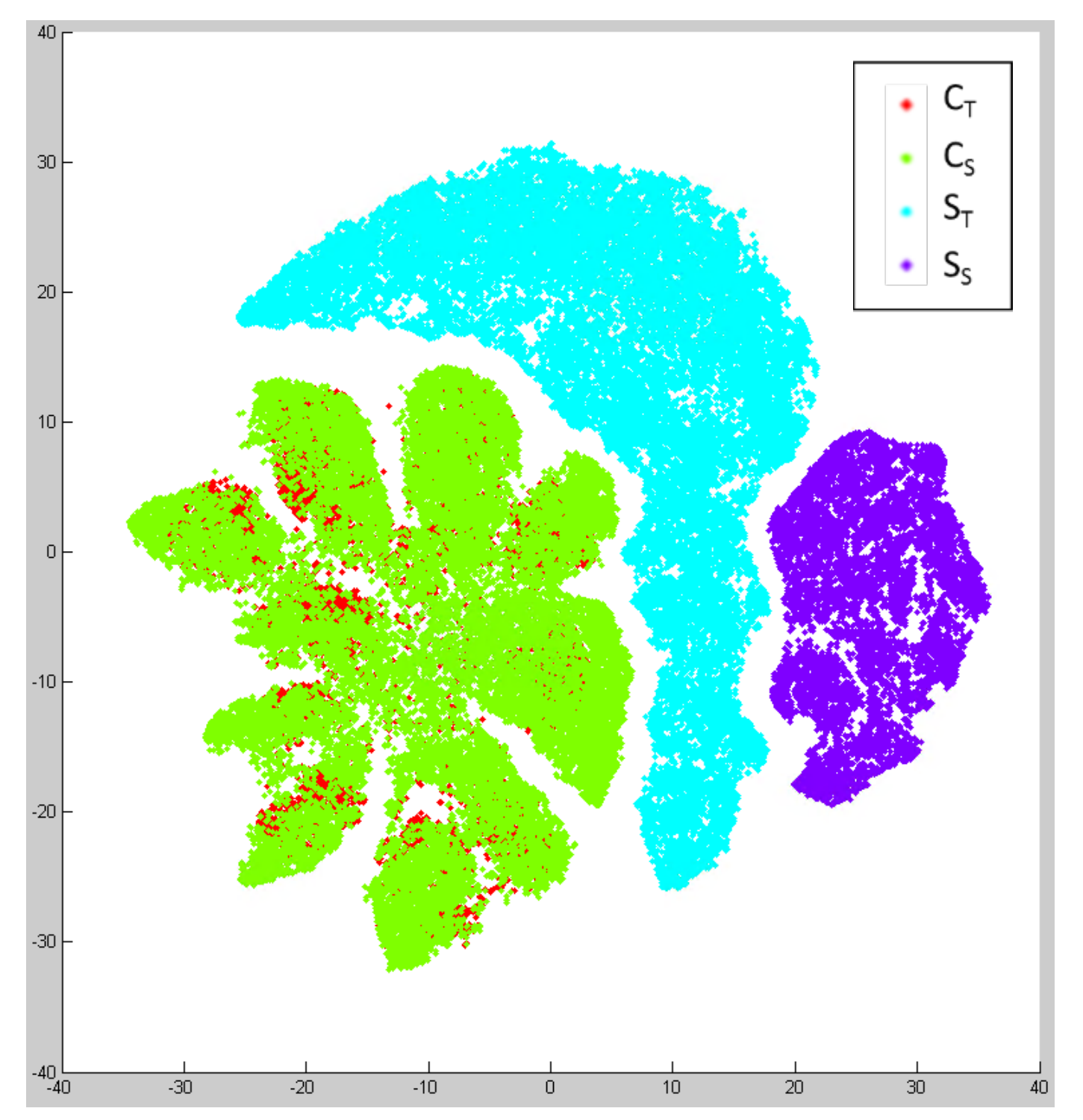}
\caption{{\bf SVHN $\rightarrow$ MNIST}}
\label{fig:fig3c}
\end{subfigure}
\caption{Feature visualization}
\label{fig:fig3}
\end{center}
\end{figure}

\subsection{Full transfer learning}
The results of our experiment are provided in Table~\ref{table:tb1}, we observe that our method performs well in all scenarios. It is worth to note that, in order to have a fair comparison, we only compare to the methods which have the simple network architecture similar to ours.

{\bf Feature visualization}. To demonstrate the distributions of our learned features, we use t-SNE \cite{30} projection. As shown in Fig.~\ref{fig:fig3}, the common features from target and source domains are matched together and grouped into 10 main clusters clearly, especially for the simple case {\bf MNIST $\rightarrow$ USPS}. This means that our framework not only can learn the common features which cannot be distinguished between domains but also have ability to match the cluster-based distributions of two domains.  On the contrary, the specific features from target and source domains are separated well and far away from common features. It means that our framework can learn the domain specific characteristics and these characteristics are different from the common features even we did not apply any difference loss such as orthogonality constraint on these features.

\begin{figure*}[t!]
\begin{subfigure}{0.48\textwidth}
\centering
\includegraphics[width=0.8\linewidth, height=0.75cm]{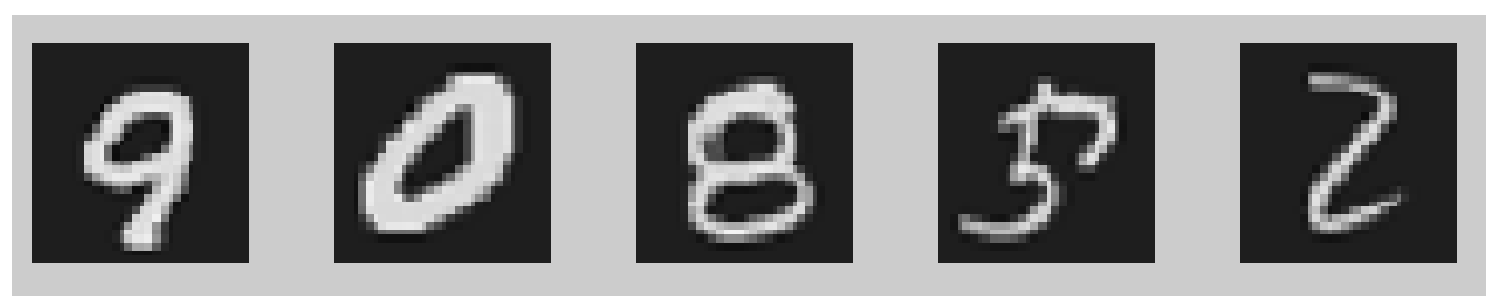}
\caption{Source image (MNIST)}
\label{fig:fig4a}
\end{subfigure}
~~~
\begin{subfigure}{0.48\textwidth}
\centering
\includegraphics[width=0.8\linewidth, height=0.75cm]{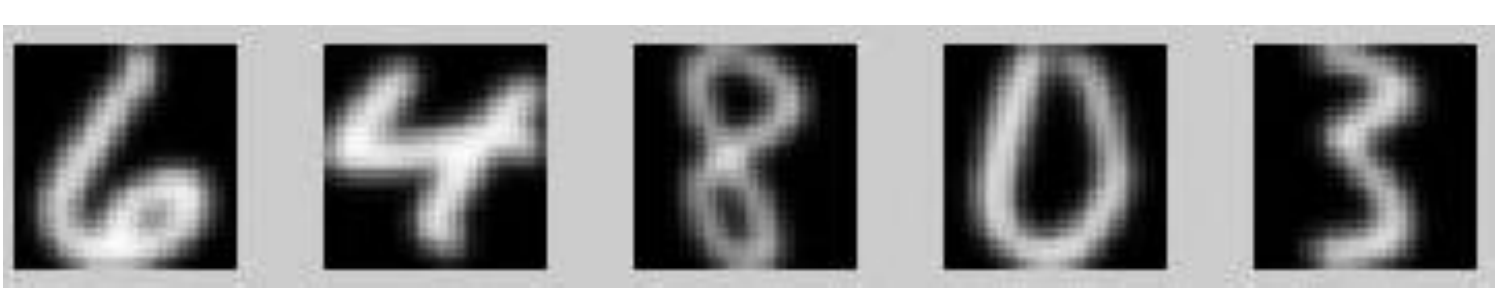}
\caption{Target image (USPS)}
\label{fig:fig4d}
\end{subfigure}
\begin{subfigure}{0.48\textwidth}
\centering
\includegraphics[width=0.8\linewidth, height=0.75cm]{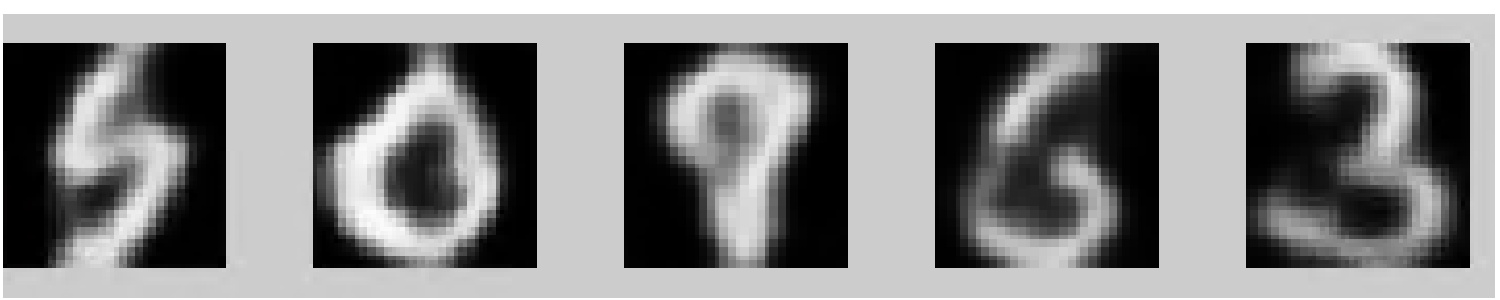}
\caption{Style-transferred images from {\bf source to target without} $L_{Sem}$}
\label{fig:fig4b}
\end{subfigure}
~~~
\begin{subfigure}{0.48\textwidth}
\centering
\includegraphics[width=0.8\linewidth, height=0.75cm]{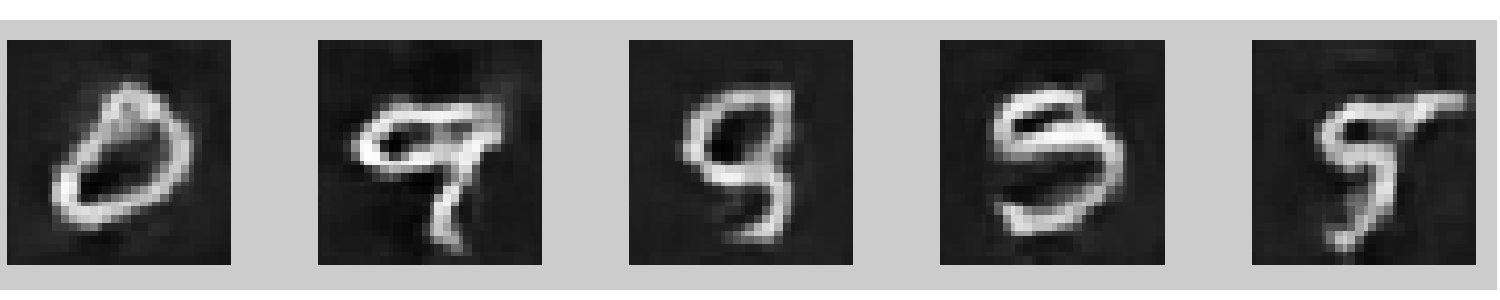}
\caption{Style-transferred images from {\bf target to source without} $L_{Sem}$}
\label{fig:fig4e}
\end{subfigure}

\begin{subfigure}{0.48\textwidth}
\centering
\includegraphics[width=0.8\linewidth, height=0.75cm]{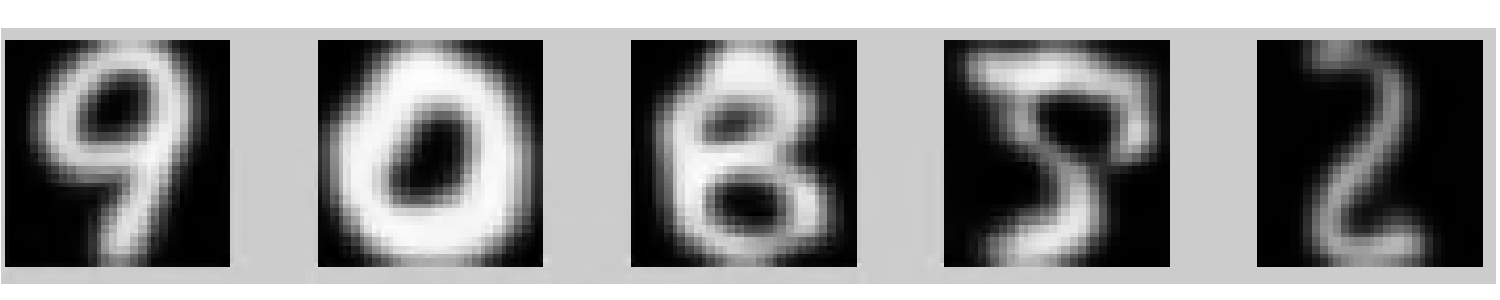}
\caption{Style-transferred images from {\bf source to target with} $L_{Sem}$}
\label{fig:fig4c}
\end{subfigure}
~~~
\begin{subfigure}{0.48\textwidth}
\centering
\includegraphics[width=0.8\linewidth, height=0.75cm]{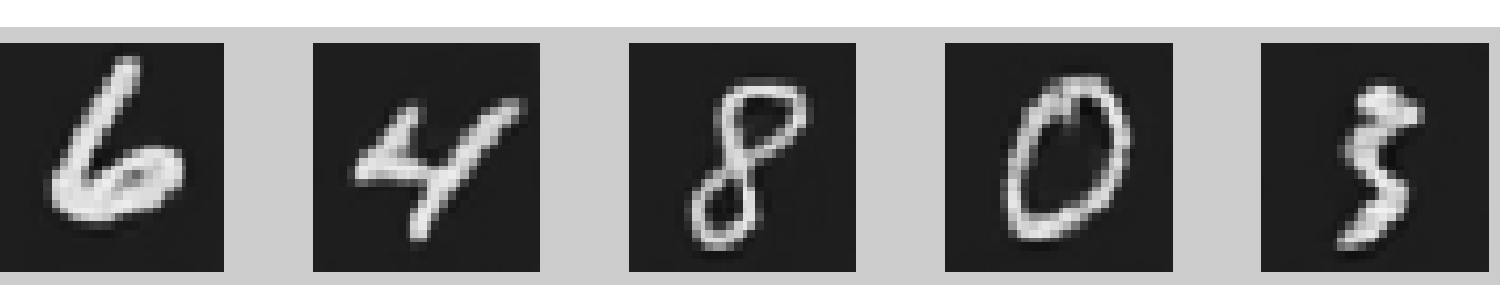}
\caption{Style-transferred images from {\bf target to source with} $L_{Sem}$}
\label{fig:fig4f}
\end{subfigure}
\caption{With and without semantic consistency loss $L_{Sem}$}
\label{fig:fig4}
\end{figure*}

{\bf With and without semantic consistency loss}. In Fig.~\ref{fig:fig4}, we show the comparison of the style-transferred images produced by the generators {\it $G_S$} and {\it $G_T$} with and without semantic consistency loss. Without semantic consistency, the style-transferred images (Fig.~\ref{fig:fig4b} and \ref{fig:fig4e}) seem to have the same style as the real ones, but they fail at preserving the semantic information. They might deliver ambiguous information to the classifier and discriminators. As shown in (Fig.~\ref{fig:fig4c} and \ref{fig:fig4f}), with semantic consistency loss, the style-transferred images successfully both preserve the semantic information and depict the style information.

\begin{figure}[t]
\begin{center}
\begin{subfigure}{0.4\textwidth}
\includegraphics[width=0.9\linewidth, height=0.75cm]{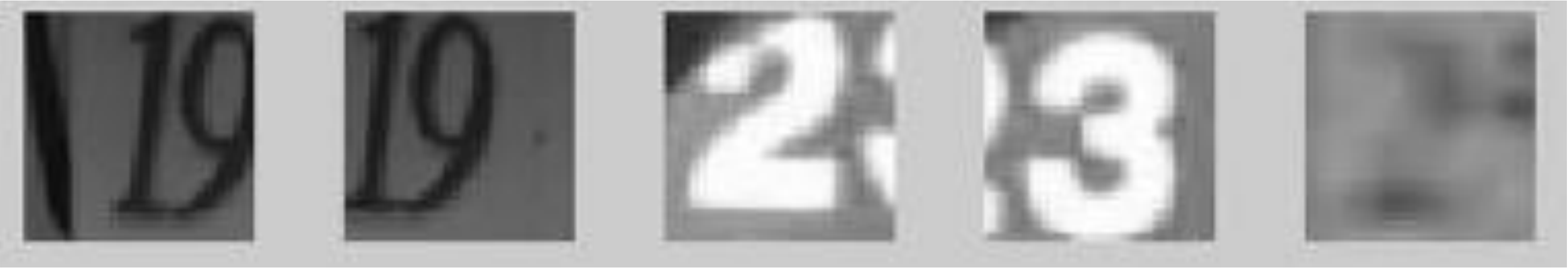}
\caption{Source image (SVHN)}
\label{fig:fig6a}
\end{subfigure}
~~~
\begin{subfigure}{0.4\textwidth}
\includegraphics[width=0.9\linewidth, height=0.75cm]{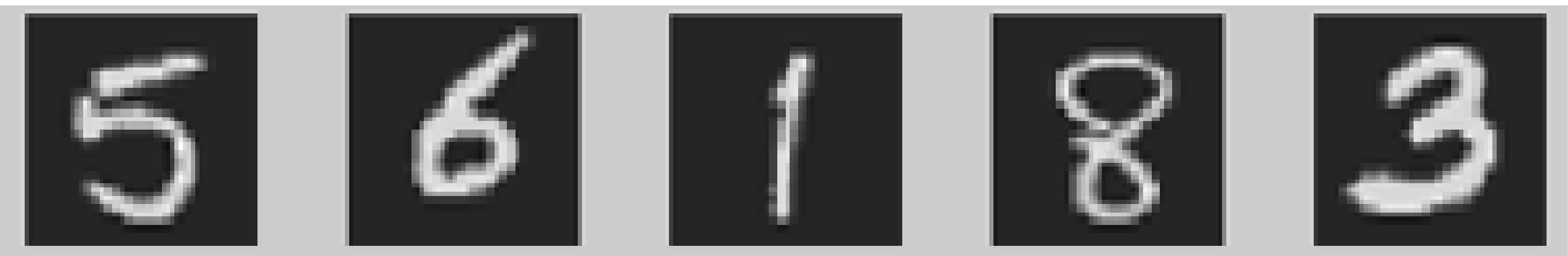}
\caption{Target image (MNIST)}
\label{fig:fig6b}
\end{subfigure}

\begin{subfigure}{0.4\textwidth}
\includegraphics[width=0.9\linewidth, height=0.75cm]{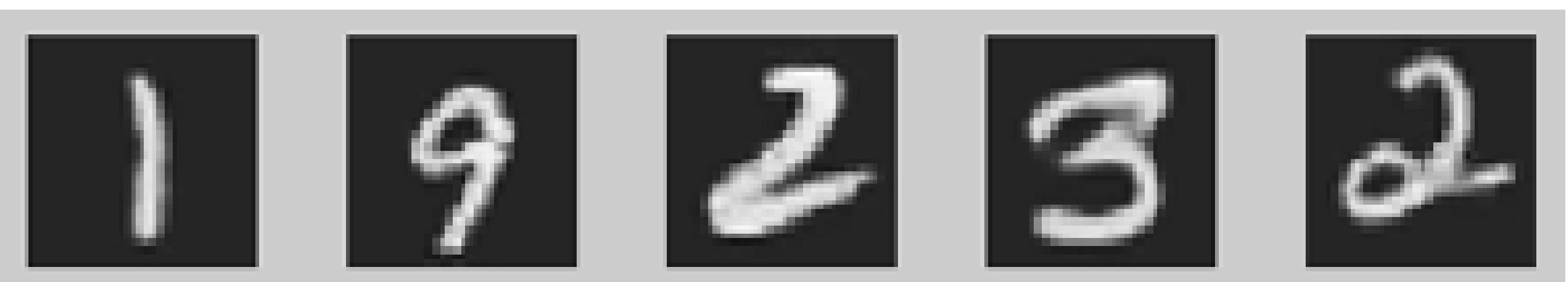}
\caption{Style-transferred images from {\bf source to target}}
\label{fig:fig6c}
\end{subfigure}
~~~
\begin{subfigure}{0.4\textwidth}
\includegraphics[width=0.9\linewidth, height=0.75cm]{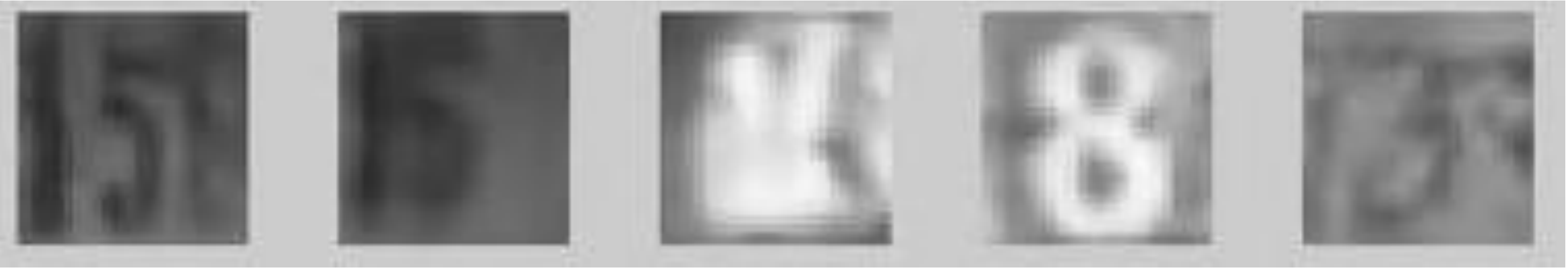}
\caption{Style-transferred images from {\bf target to source}}
\label{fig:fig6d}
\end{subfigure}
\begin{subfigure}{0.4\textwidth}
\includegraphics[width=0.9\linewidth, height=0.75cm]{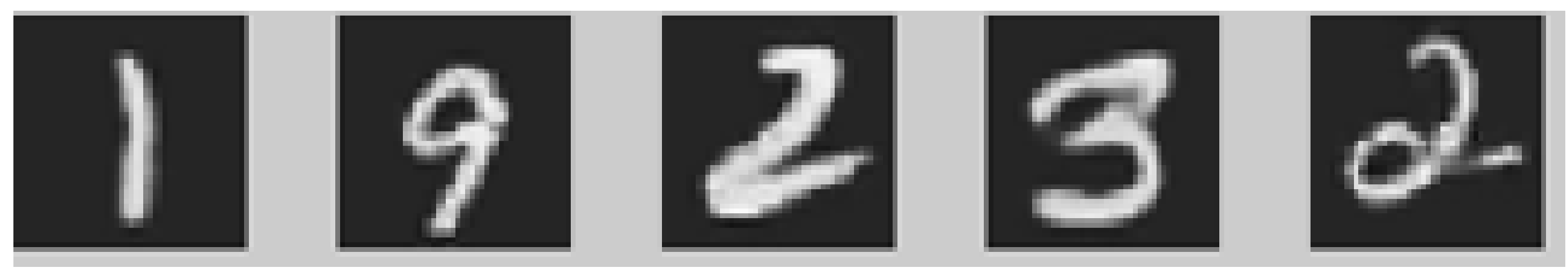}
\caption{Source's common part}
\label{fig:fig6e}
\end{subfigure}
~~~
\begin{subfigure}{0.4\textwidth}
\includegraphics[width=0.9\linewidth, height=0.75cm]{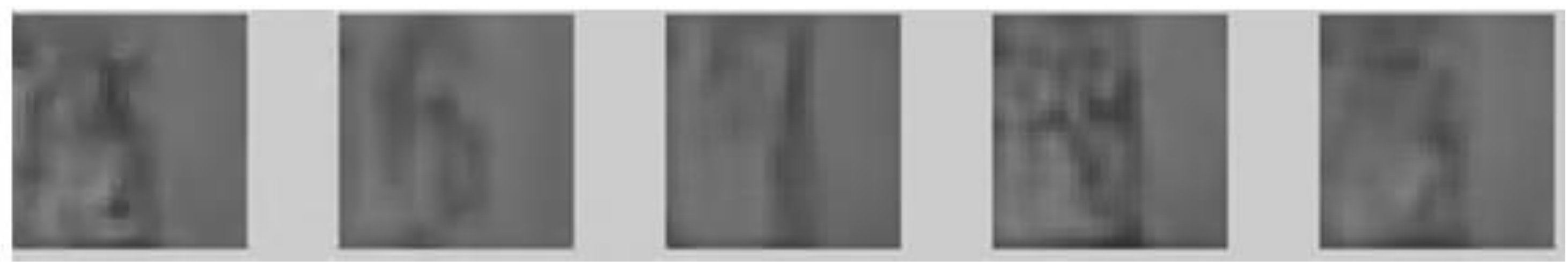}
\caption{Target's common part}
\label{fig:fig6f}
\end{subfigure}
\begin{subfigure}{0.4\textwidth}
\includegraphics[width=0.9\linewidth, height=0.75cm]{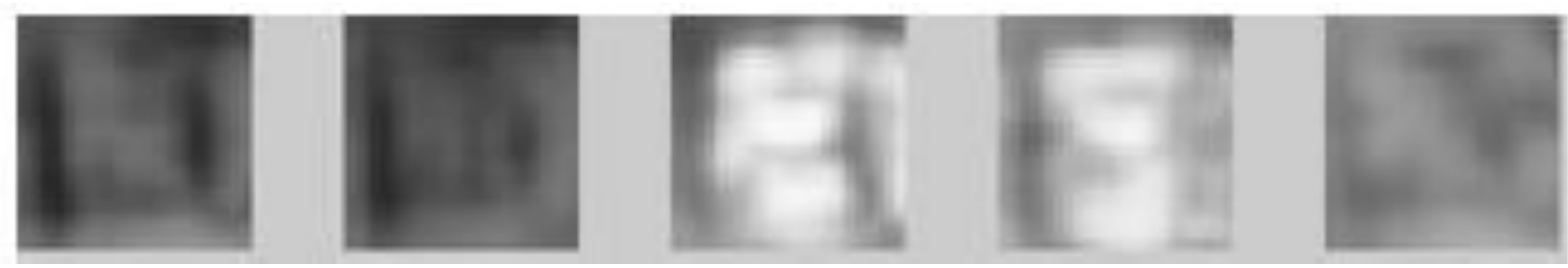}
\caption{Source's specific part}
\label{fig:fig6g}
\end{subfigure}
~~~
\begin{subfigure}{0.4\textwidth}
\includegraphics[width=0.9\linewidth, height=0.75cm]{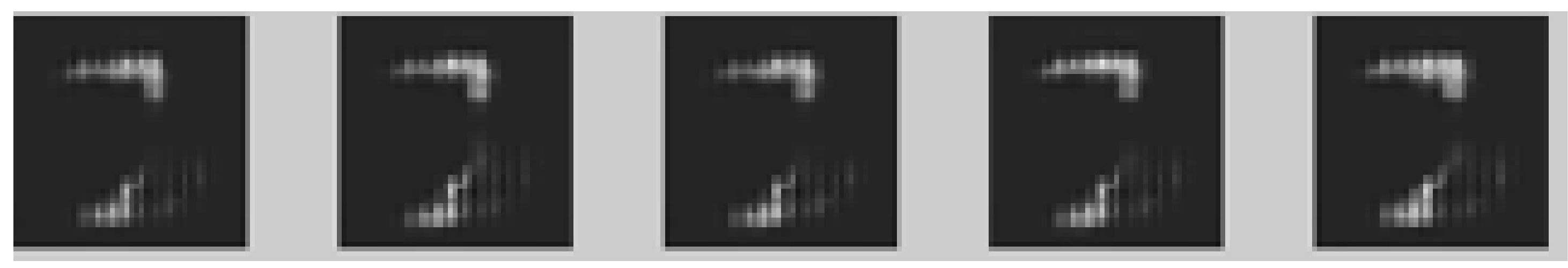}
\caption{Target's specific part}
\label{fig:fig6h}
\end{subfigure}
\caption{SVHN $\rightarrow$ MNIST}
\label{fig:fig6}
\end{center}
\end{figure}

{\bf Visualizing the common and specific parts in image domain}. In Fig.~\ref{fig:fig6} we show some examples for {\bf SVHN $\rightarrow$ MNIST} case. We also try to visualize the common and specific parts in image domain by inhibiting the remaining part before inputting to the Generators. For example, if we want to visualize the common parts, we will set all specific parts equal to zero then concatenate them and input into the corresponding Generator. As shown in the Fig.~\ref{fig:fig6e} and \ref{fig:fig6f}, the common parts of each domain only store the information of ``digits''. Meanwhile, the specific parts encode the style information such as the contrast, color, size, ... as shown clearly in Fig.~\ref{fig:fig6g}. For the target's specific parts shown in Fig.~\ref{fig:fig6h}, all the images are almost same because the style of all the images in target domain (MNIST) are same. This experiment partly demonstrates our aforementioned statements about our framework's abilities to learn disentangled representations and transfer the style across domain.

The visualizing results for Syn-Signs to GTSRB are given in the Supplemental Material D.

\begin{figure}[h]
\centering
\begin{subfigure}{0.48\textwidth}
\centering
\includegraphics[width=0.8\linewidth, height=4cm]{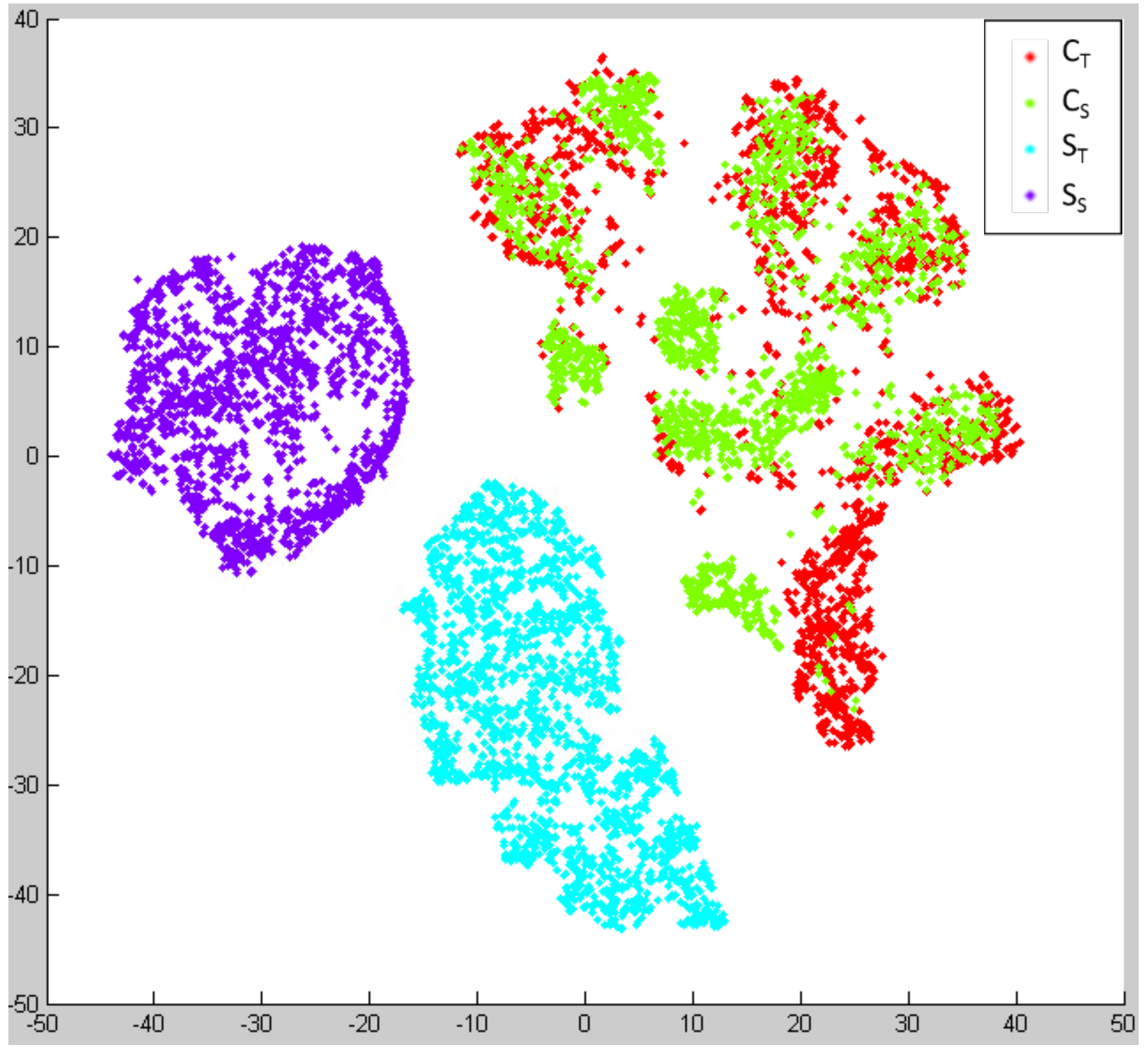}
\caption{Without {\it $L_{feedback}$} and {\it $L_{Entropy}$}}
\label{fig:fig5a}
\end{subfigure}
\begin{subfigure}{0.48\textwidth}
\centering
\includegraphics[width=0.8\linewidth, height=4cm]{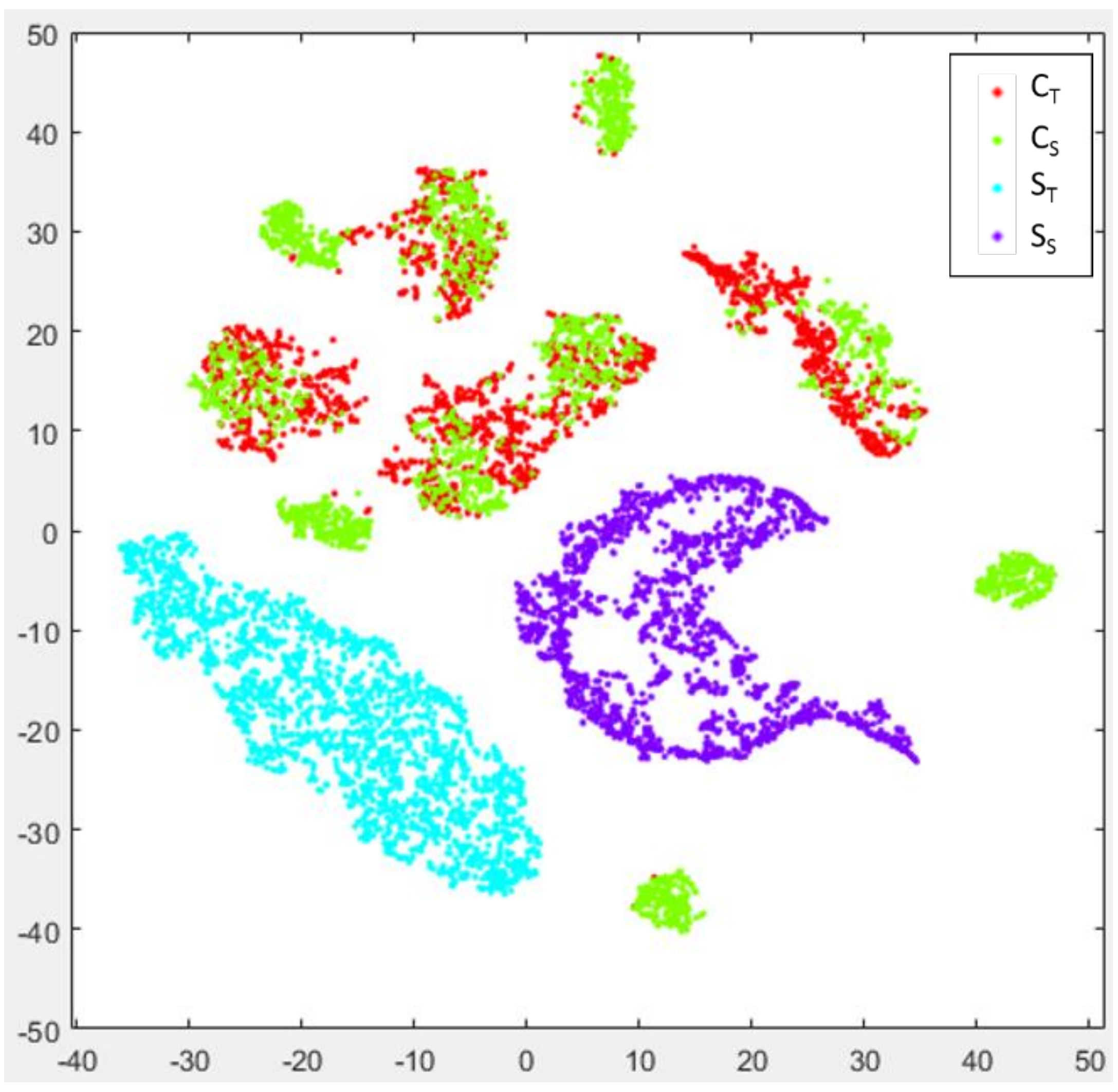}
\caption{Without {\it $L_{Entropy}$}}
\label{fig:fig5b}
\end{subfigure}
\begin{subfigure}{0.48\textwidth}
\centering
\includegraphics[width=0.8\linewidth, height=4cm]{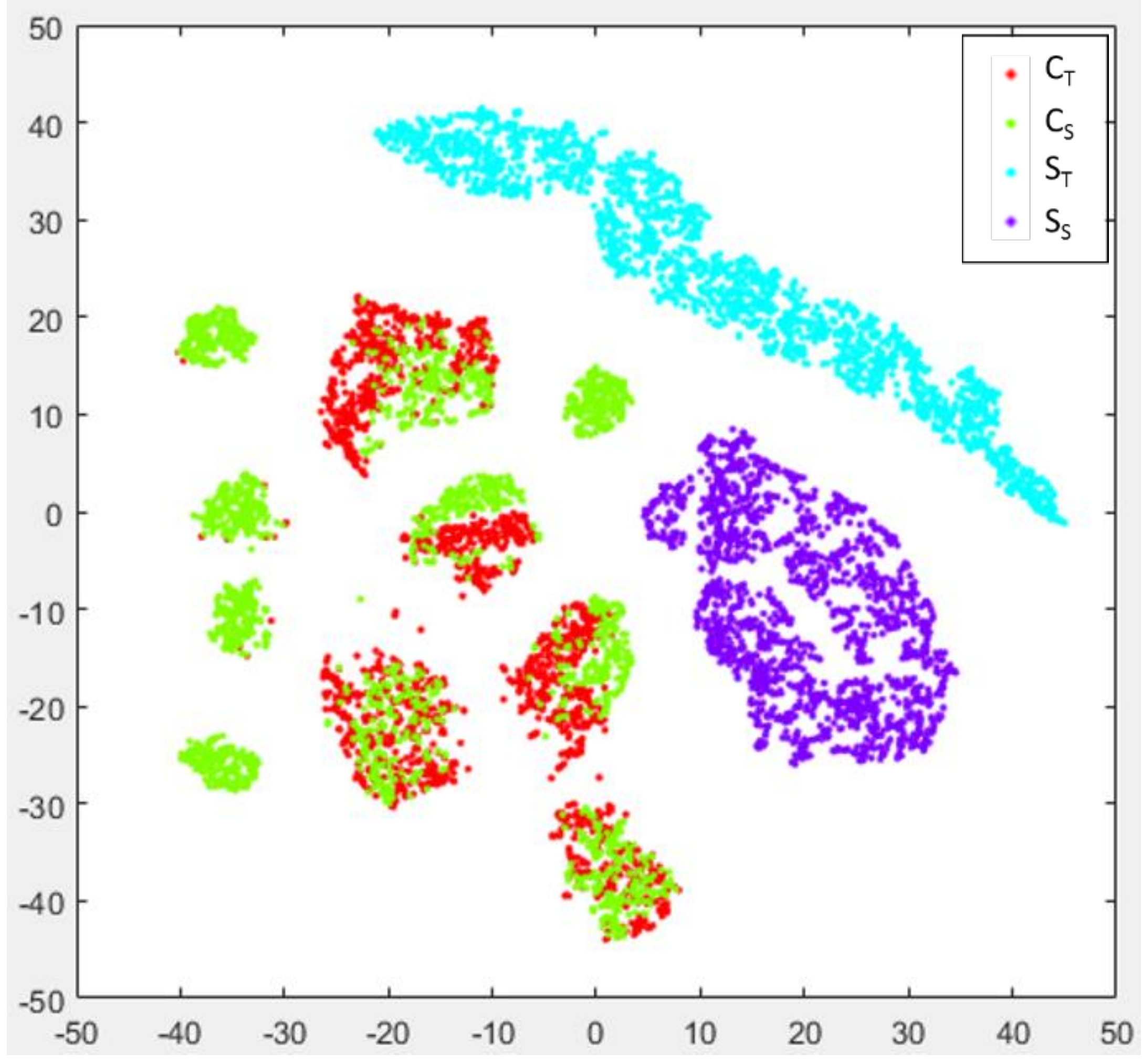}
\caption{Proposed method}
\label{fig:fig5c}
\end{subfigure}
\caption{Visualization of learned features when partial transferring from MNIST to USPS.}
\label{fig:fig5}
\end{figure}

\setlength{\tabcolsep}{4pt}
\begin{table}[!]
\begin{center}
\caption{Experimental results on partial transfer learning between MNIST and USPS}
\label{table:tb2}
\begin{tabular}{lc}
\hline\noalign{\smallskip}
 & {\bf MNIST $\rightarrow$ USPS}\\
\noalign{\smallskip}
\hline
\noalign{\smallskip}
Without {\it $L_{feedback}$} and {\it $L_{Entropy}$} & 75.22\% \\
Without {\it $L_{Entropy}$} & 89.50\% \\
\hline
Proposed method	& {\bf 94.78\%} \\
\hline
\end{tabular}
\end{center}
\end{table}
\setlength{\tabcolsep}{1.4pt}

\subsection{Partial Transfer Learning}
In order to quickly test whether our framework is capable of solving partial transfer learning problem, we still use the digit datasets and randomly select 5 classes to form the target domain data (USPS), and apply our framework with 3 different settings: (1) without feedback and entropy minimization losses, (2) without entropy minimization losses and (3) proposed method. Based on the results in Table~\ref{table:tb2} and the t-SNE visualization in Fig.~\ref{fig:fig5}, we can understand the importance of {\it $L_{feedback}$} and {\it $L_{Entropy}$} in our framework especially for partial transfer learning case. As aforementioned, by enhancing the harmony and the stability of extractors and generators, {\it $L_{feedback}$} improves the performance significantly by {\bf 14.28\%}. Without {\it $L_{feedback}$}, the outlier source clusters seem to cause certain difficulties to transferring process, as shown in Fig.~\ref{fig:fig5a}, the corresponding common clusters between source and target are not aligned well and some clusters are mismatched. With {\it $L_{feedback}$}, as shown in Fig.~\ref{fig:fig5b}, the corresponding classes between source and target are matched, however it still has ambiguous areas between classes. In Fig.~\ref{fig:fig5c}, we can recognize that {\it $L_{Entropy}$} solves this problem and boosts the performance ({\bf 5.28\%}) by clustering semantic classes together and thereby increasing the distance between clusters and then reducing the ambiguity. Moreover, comparing the performance and the feature distribution of our method for two transferring cases (full transferring and partial transferring), we can partly assert that our network can solve the partial transferring problem efficiently.



\section{Conclusions}

In this paper, we proposed a unified framework for unsupervised domain adaptation that can disentangle the feature domain into two main parts: common features across domains, and specific features of each domain. The common features are used to embed the content information that are useful for classification across domains, and the specific features are used to encode the domain specific characteristics. To enhance the transferability of common features, we proposed the novel idea for feature exchange across domains which also allows us to embed style transfer function into our framework. Besides, we also introduced the feedback design and semantic consistency loss to improve the harmony and the stability of the components in the framework. Last but not least, entropy minimization losses were applied in our framework as a refining method to encourage the self-clustering ability in the target domain and thereby make the transfer process easier. The experiments confirm the improvement of our network in solving both of full transfer problem and partial transfer problem. Besides, the results also show the potential of our network in image style transfer.

\bibliographystyle{splncs}
\bibliography{egbib}
\clearpage

\section{Supplementary Material}
\subsection{Datasets}
\label{sect:a}

The datasets used in this paper are described in Table ~\ref{table:tb13}. It should be noted that for GTSRB dataset, the original number of training data is 39,209 and testing data is 12,630. However, the label for testing data is not available, we randomly select 31,367 samples from original training set to form the new training set and the rest for evaluation. In addition to the datasets listed in Table ~\ref{table:tb13}, there are other datasets for domain adaptation such as OFFICE and Office+Caltech. Due to high image variations and small amounts of data, some works pre-train their models via other big datasets before performing domain adaptation in these datasets. However, by using other pre-trained models, it may not reveal pure message transfer between two domains. Thus, we insist to train our network from scratch. For fair comparison, these datasets are not included in our experiment now.

\setlength{\tabcolsep}{4pt}
\begin{table}[h]
\begin{center}
\caption{Datasets}
\label{table:tb13}
\begin{tabular}{p{2 cm} | p{1.5cm} | p{1.5cm} | p{1.5cm} | p{1.5cm} | p{1.5cm}}
\hline\noalign{\smallskip}
 						& \#train 						&\#test					&\#classes		&Resolution		&Channels\\
\noalign{\smallskip}
\hline
\noalign{\smallskip}
USPS 					& 7,291 						& 2,007  					& 10				&16 x 16			&Gray		\\
MNIST 				& 60,000 						& 10,000 					& 10				&28 x 28			&Gray		\\
SVHN 					& 73,257 						& 26,032 					& 10				&32 x 32			&RGB		\\
Syn-Signs 			& 100,000 					& - 						& 43				&40 x 40			&RGB		\\
GTSRB				& 31,367 						& 7,842 					& 43				&varies			&RGB		\\
\end{tabular}
\end{center}
\end{table}

\subsection{Network Architectures}
\label{sect:b}
Our network architectures are shown in \Cref{table:tb14,table:tb3,table:tb4,table:tb5,table:tb6,table:tb7,table:tb8,table:tb9,table:tb10,table:tb11,table:tb12}.

\setlength{\tabcolsep}{4pt}
\begin{table}[p]
\begin{center}
\caption{Feature Extraction Architecture for MNIST to USPS \& SVHN to MNIST}
\label{table:tb14}
\begin{tabular}{p{4.5cm} | p{5cm}}
\hline\noalign{\smallskip}
 						Description 									&Shape									\\
\noalign{\smallskip}
\hline
\noalign{\smallskip}
						32 x 32 gray image 							& 32 x 32 x 1  							\\
						Conv 5 x 5 x 64, pad 2, ReLU				& 32 x 32 x 64 							\\
						Max-pool, 2 x 2 								& 16 x 16 x 64  							\\
						Conv 5 x 5 x 128, ReLU 					& 12 x 12 x 128 							\\
						Max-pool, 2 x 2 								& 6 x 6 x 128 							\\
						Conv 5 x 5 x 256, ReLU 					& 2 x 2 x 256 							\\
						Max-pool, 2 x 2 								& 1 x 1 x 256 							\\
						Fully connected, 256 units 					& 256 (Specific: 128, Common: 128) 	\\
\end{tabular}
\end{center}
\end{table}

\setlength{\tabcolsep}{4pt}
\begin{table}[h]
\begin{center}
\caption{Feature Extraction Architecture for Syn-Signs to GTSRB}
\label{table:tb3}
\begin{tabular}{p{5cm} | p{6cm}}
\hline\noalign{\smallskip}
 						Description 									&Shape											\\
\noalign{\smallskip}
\hline
\noalign{\smallskip}
						40 x 40 RGB image 							& 40 x 40 x 3  									\\
						Conv 5 x 5 x 128, batch norm, ELU		& 36 x 36 x 128 									\\
						Max-pool, 2 x 2 								& 18 x 18 x 128  								\\
						Conv 3 x 3 x 256, batch norm, ELU		& 16 x 16 x 256 									\\
						Max-pool, 2 x 2 								& 8 x 8 x 256 									\\
						Conv 5 x 5 x 512, batch norm, ELU		& 4 x 4 x 512 									\\
						Max-pool, 2 x 2 								& 2 x 2 x 512 (Specific: 256, Common: 256)	\\
\end{tabular}
\end{center}
\end{table}

\setlength{\tabcolsep}{4pt}
\begin{table}[h]
\begin{center}
\caption{Generator Architecture for MNIST to USPS}
\label{table:tb4}
\begin{tabular}{p{4.5cm} | p{2cm}}
\hline\noalign{\smallskip}
 						Description 									&Shape									\\
\noalign{\smallskip}
\hline
\noalign{\smallskip}
						Specific: 128, Common: 128 				& 256  									\\
						Unpool, 2 x 2									& 2 x 2 x 256 							\\
						Deconv 5 x 5 x 512, ReLU 					& 6 x 6 x 512  							\\
						Unpool, 2 x 2 								& 12 x 12 x 512 							\\
						Deconv 5 x 5 x 256, ReLU 					& 16 x 16 x 256 							\\
						Unpool, 2 x 2 								& 32 x 32 x 256 							\\
						Deconv 5 x 5 x 128, pad 2, ReLU			& 32 x 32 x 128 							\\
						Deconv 5 x 5 x 1, pad 2, Sigmoid			& 32 x 32 x 1 							\\

\end{tabular}
\end{center}
\end{table}

\setlength{\tabcolsep}{4pt}
\begin{table}[h]
\begin{center}
\caption{Generator Architecture for SVHN to MNIST}
\label{table:tb5}
\begin{tabular}{p{4.5cm} | p{2cm}}
\hline\noalign{\smallskip}
 						Description 									&Shape									\\
\noalign{\smallskip}
\hline
\noalign{\smallskip}
						Specific: 128, Common: 128 				& 256  									\\
						Fully connected, 256 units, ReLU 			& 256 										\\
						Unpool, 2 x 2									& 2 x 2 x 256 							\\
						Deconv 5 x 5 x 512, ReLU 					& 6 x 6 x 512  							\\
						Unpool, 2 x 2 								& 12 x 12 x 512 							\\
						Deconv 5 x 5 x 256, ReLU 					& 16 x 16 x 256 							\\
						Unpool, 2 x 2 								& 32 x 32 x 256 							\\
						Deconv 5 x 5 x 128, pad 2, ReLU			& 32 x 32 x 128 							\\
						Deconv 5 x 5 x 1, pad 2, Sigmoid			& 32 x 32 x 1 							\\

\end{tabular}
\end{center}
\end{table}

\setlength{\tabcolsep}{4pt}
\begin{table}[h]
\begin{center}
\caption{Generator Architecture for Syn-Signs to GTSRB}
\label{table:tb6}
\begin{tabular}{p{6cm} | p{2cm}}
\hline\noalign{\smallskip}
 						Description 											&Shape									\\
\noalign{\smallskip}
\hline
\noalign{\smallskip}
						Specific: 2 x 2 x 256, Common: 2 x 2 x 256 		& 2 x 2 x 512  							\\
						Unpool, 2 x 2											& 4 x 4 x 512 							\\
						Deconv 5 x 5 x 128, batch norm, ELU 			& 8 x 8 x 128  							\\
						Unpool, 2 x 2 										& 16 x 16 x 128 							\\
						Deconv 5 x 5 x 32, batch norm, ELU 				& 20 x 20 x 32 							\\
						Unpool, 2 x 2 										& 40 x 40 x 32 							\\
						Deconv 3 x 3 x 16, pad 1, batch norm, ELU		& 40 x 40 x 16 							\\
						Deconv 3 x 3 x 3, pad 1, Sigmoid					& 40 x 40 x 3 							\\

\end{tabular}
\end{center}
\end{table}

\setlength{\tabcolsep}{4pt}
\begin{table}[h]
\begin{center}
\caption{Feature Discriminator Architecture for MNIST to USPS \& SVHN to MNIST}
\label{table:tb7}
\begin{tabular}{p{5cm} | p{4cm}}
\hline\noalign{\smallskip}
 						Description 									&Shape									\\
\noalign{\smallskip}
\hline
\noalign{\smallskip}
						Common: 128 								& 128  									\\
						Fully connected, 128 units, ReLU			& 128 										\\
						Fully connected, 128 units, ReLU			& 128 										\\
						Fully connected, 11 units 					& 11 (Classes: 10, Domain: 1)			\\
\end{tabular}
\end{center}
\end{table}

\setlength{\tabcolsep}{4pt}
\begin{table}[h]
\begin{center}
\caption{Feature Discriminator Architecture for Syn-Signs to GTSRB}
\label{table:tb8}
\begin{tabular}{p{6.5cm} | p{4cm}}
\hline\noalign{\smallskip}
 						Description 											&Shape									\\
\noalign{\smallskip}
\hline
\noalign{\smallskip}
						Common: 2 x 2 x 256 								& 2 x 2 x 256								\\
						Fully connected, 512 units, batch norm, ELU		& 512 										\\
						Fully connected, 256 units, batch norm, ELU		& 256 										\\
						Fully connected, 44 units 							& 44 (Classes: 43, Domain: 1)			\\
\end{tabular}
\end{center}
\end{table}

\setlength{\tabcolsep}{4pt}
\begin{table}[h]
\begin{center}
\caption{Image Discriminator Architecture for MNIST to USPS \& SVHN to MNIST}
\label{table:tb9}
\begin{tabular}{p{5cm} | p{5cm}}
\hline\noalign{\smallskip}
 						Description 									&Shape									\\
\noalign{\smallskip}
\hline
\noalign{\smallskip}
						32 x 32 gray image 							& 32 x 32 x 1  							\\
						Conv 5 x 5 x 64, pad 2, ReLU				& 32 x 32 x 64 							\\
						Max-pool, 2 x 2 								& 16 x 16 x 64  							\\
						Conv 5 x 5 x 128, ReLU 					& 12 x 12 x 128 							\\
						Max-pool, 2 x 2 								& 6 x 6 x 128 							\\
						Conv 5 x 5 x 256, ReLU 					& 2 x 2 x 256 							\\
						Max-pool, 2 x 2 								& 1 x 1 x 256 							\\
						Fully connected, 128 units, ReLU 			& 128										\\
						Fully connected, 11 units 					& 11 (Classes: 10, Real/Fake: 1)			\\
\end{tabular}
\end{center}
\end{table}

\setlength{\tabcolsep}{4pt}
\begin{table}[h]
\begin{center}
\caption{Image Discriminator Architecture for Syn-Signs to GTSRB}
\label{table:tb10}
\begin{tabular}{p{6.5cm} | p{4.2cm}}
\hline\noalign{\smallskip}
 						Description 											&Shape											\\
\noalign{\smallskip}
\hline
\noalign{\smallskip}
						40 x 40 RGB image 									& 40 x 40 x 3  									\\
						Conv 5 x 5 x 128, batch norm, ELU				& 36 x 36 x 128 									\\
						Max-pool, 2 x 2 										& 18 x 18 x 128  								\\
						Conv 3 x 3 x 256, batch norm, ELU				& 16 x 16 x 256 									\\
						Max-pool, 2 x 2 										& 8 x 8 x 256 									\\
						Conv 5 x 5 x 512, batch norm, ELU				& 4 x 4 x 512 									\\
						Max-pool, 2 x 2 										& 2 x 2 x 512 									\\
						Fully connected, 512 units, batch norm, ELU		& 512 												\\
						Fully connected, 256 units, batch norm, ELU		& 256 												\\
						Fully connected, 44 units 							& 44 (Classes: 43, Real/Fake: 1)				\\
\end{tabular}
\end{center}
\end{table}

\setlength{\tabcolsep}{4pt}
\begin{table}[h]
\begin{center}
\caption{Classifier Architecture for MNIST to USPS \& SVHN to MNIST}
\label{table:tb11}
\begin{tabular}{p{5cm} | p{4cm}}
\hline\noalign{\smallskip}
 						Description 									&Shape									\\
\noalign{\smallskip}
\hline
\noalign{\smallskip}
						Common: 128 								& 128  									\\
						Fully connected, 128 units, ReLU			& 128 										\\
						Fully connected, 128 units, ReLU			& 128 										\\
						Fully connected, 10 units 					& 10 										\\
\end{tabular}
\end{center}
\end{table}

\setlength{\tabcolsep}{4pt}
\begin{table}[h]
\begin{center}
\caption{Classifier Architecture for Syn-Signs to GTSRB}
\label{table:tb12}
\begin{tabular}{p{6.5cm} | p{4cm}}
\hline\noalign{\smallskip}
 						Description 											&Shape									\\
\noalign{\smallskip}
\hline
\noalign{\smallskip}
						Common: 2 x 2 x 256 								& 2 x 2 x 256								\\
						Fully connected, 512 units, batch norm, ELU		& 512 										\\
						Fully connected, 256 units, batch norm, ELU		& 256 										\\
						Fully connected, 43 units 							& 43 										\\
\end{tabular}
\end{center}
\end{table}

\subsection{Training Procedures and Hyperparameters}
\label{sec:c}
Because we would like to train our network from scratch, before training the whole network, we train each part of it one by one by the following steps:
\begin{itemize}
\item {\bf Step 1}: Train a good classifier and feature extraction using source data with the learning rate 0.01. (20,000 iteration for MNIST to USPS, 43,000 iteration for SVHN to MNIST, and 50,000 iteration for Syn-Signs to GTSRB)
\item {\bf Step 2}: Add the generator into the framework and continue train it using reconstruction loss (for both source and target data) and classification loss (for source data only) with the learning rate 0.01. (5,000 iteration for MNIST to USPS, 16,000 iteration for SVHN to MNIST, and 7,000 iteration for Syn-Signs to GTSRB)
\item {\bf Step 3}: Duplicate the network (one for source and one for target) and add the feature discriminator into the framework, continue train it using {\it $L_{adv\_fea}$} (loss\_weight: 1), reconstruction loss (loss\_weight: 1) and classification loss (loss\_weight: 10) with the learning rate 0.001. (25,000 iteration for MNIST to USPS, 25,000 iteration for SVHN to MNIST, and 27,000 iteration for Syn-Signs to GTSRB)
\item {\bf Step 4}: Add the image discriminators into the framework and initilize these discriminators using trained generator's weights, continue train the framework using {\it $L_{adv\_img}$} (loss\_weight: 1), {\it $L_{adv\_fea}$} (loss\_weight: 1), reconstruction loss (loss\_weight: 1) and classification loss (loss\_weight: 10) with the learning rate 0.001. (5,000 iteration for MNIST to USPS, 7,000 iteration for SVHN to MNIST, and 9,000 iteration for Syn-Signs to GTSRB)
\end{itemize}

When we achieve the initial weights for the whole parts of our network by the previous process, we start to train it end-to-end by the following steps:
\begin{itemize}
\item {\bf Step 1}: We first warm-up the whole network by continue training it without {\bf{\emph{$L_{Entropy}$}}} with the learning rate 0.001. (39,000 iterations for MNIST to USPS, 129,500 iteration for SVHN to MNIST, and 36,000 iteration for Syn-Signs to GTSRB)
\item {\bf Step 2}: We copy (replace) target network’s weights by the trained source network’s weights and then re-train without {\bf{\emph{$L_{Entropy}$}}} with the learning rate 0.001. (20,000 iterations for MNIST to USPS, 39,600 iteration for SVHN to MNIST, and 25,600 iteration for Syn-Signs to GTSRB)
\item {\bf Step 3}: We refine the network by training with {\bf{\emph{$L_{Entropy}$}}} loss and learning rate 0.0001. (2,000 iterations for MNIST to USPS, 16,000 iteration for SVHN to MNIST, and 6,000 iteration for Syn-Signs to GTSRB)
\end{itemize}

During training the whole network, we can repeat step 2 and step 3 to achieve better initialization. As the network converges, we can also reduce the learning rate by ten times and continue training to get higher performance.

\subsection{Visualizing Results for Syn-Signs $\rightarrow$ GTSRB}
\label{sec:d}

\begin{figure}[t]
\begin{center}
\begin{subfigure}{0.47\textwidth}
\includegraphics[width=0.95\linewidth, height=0.95cm]{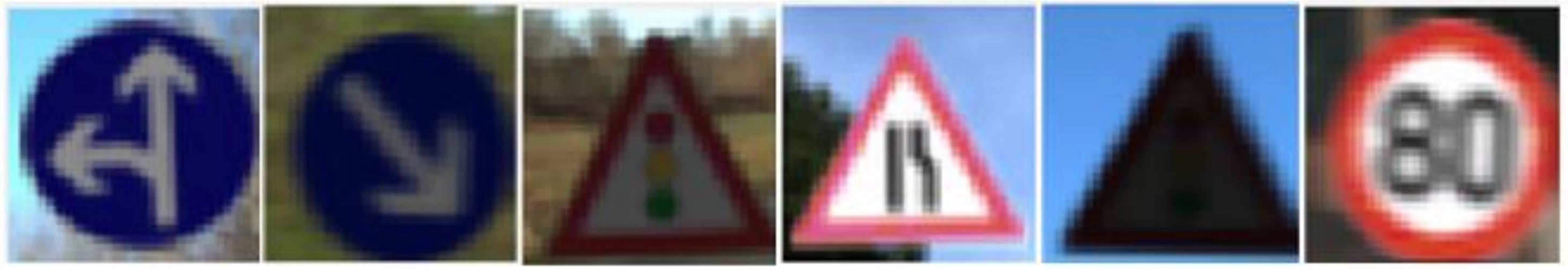}
\caption{Source image (Syn-Signs)}
\label{fig:fig11a}
\end{subfigure}
~~~
\begin{subfigure}{0.47\textwidth}
\includegraphics[width=0.95\linewidth, height=0.95cm]{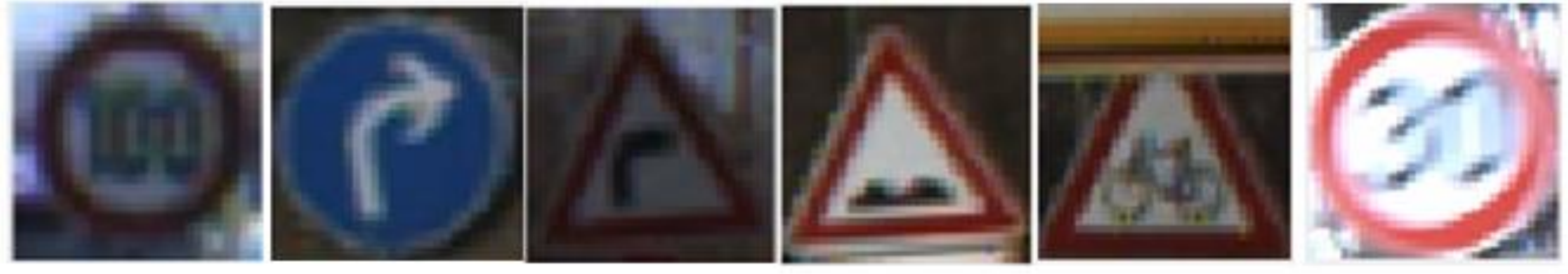}
\caption{Target image (GTSRB)}
\label{fig:fig11b}
\end{subfigure}

\begin{subfigure}{0.47\textwidth}
\includegraphics[width=0.95\linewidth, height=0.95cm]{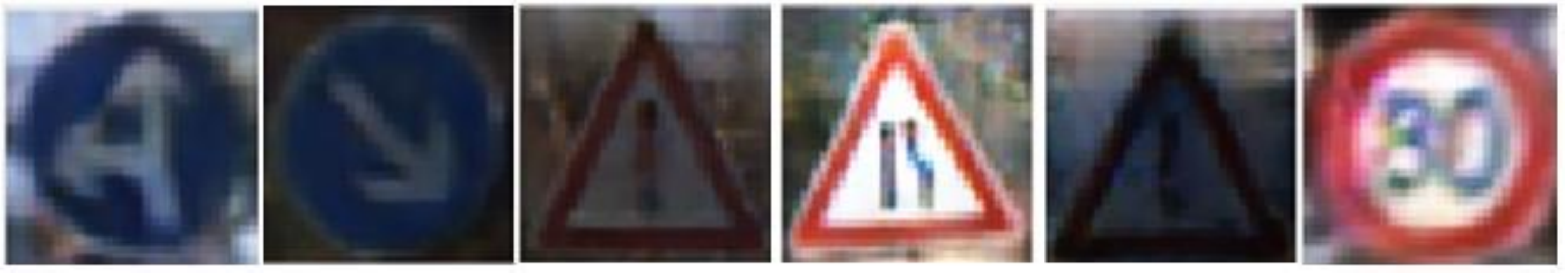}
\caption{Style-transferred images from {\bf source to target}}
\label{fig:fig11c}
\end{subfigure}
~~~
\begin{subfigure}{0.47\textwidth}
\includegraphics[width=0.95\linewidth, height=0.95cm]{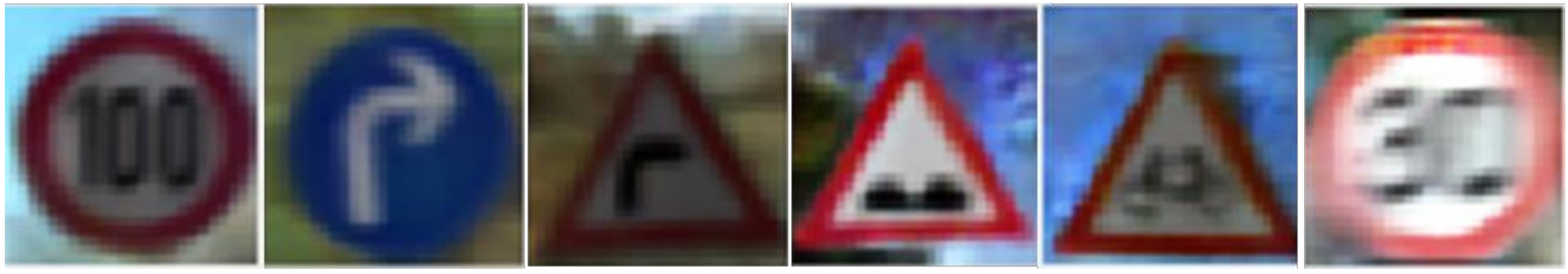}
\caption{Style-transferred images from {\bf target to source}}
\label{fig:fig11d}
\end{subfigure}
\begin{subfigure}{0.47\textwidth}
\includegraphics[width=0.95\linewidth, height=0.95cm]{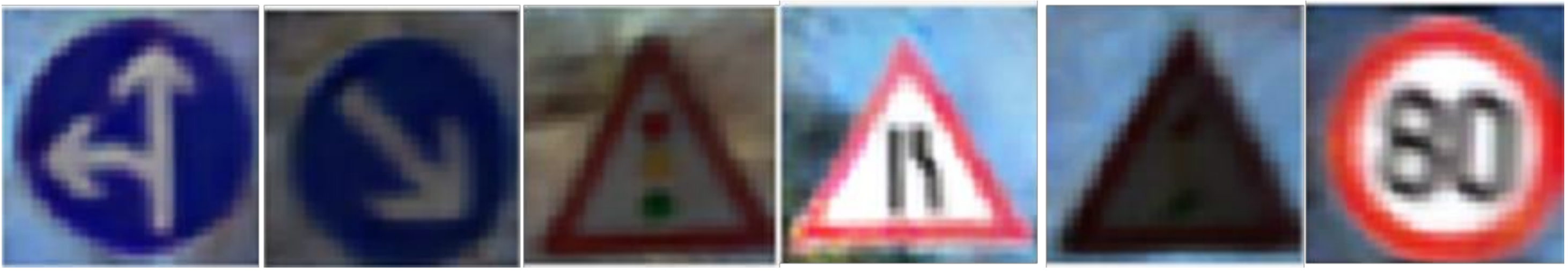}
\caption{Source's common part}
\label{fig:fig11e}
\end{subfigure}
~~~
\begin{subfigure}{0.47\textwidth}
\includegraphics[width=0.95\linewidth, height=0.95cm]{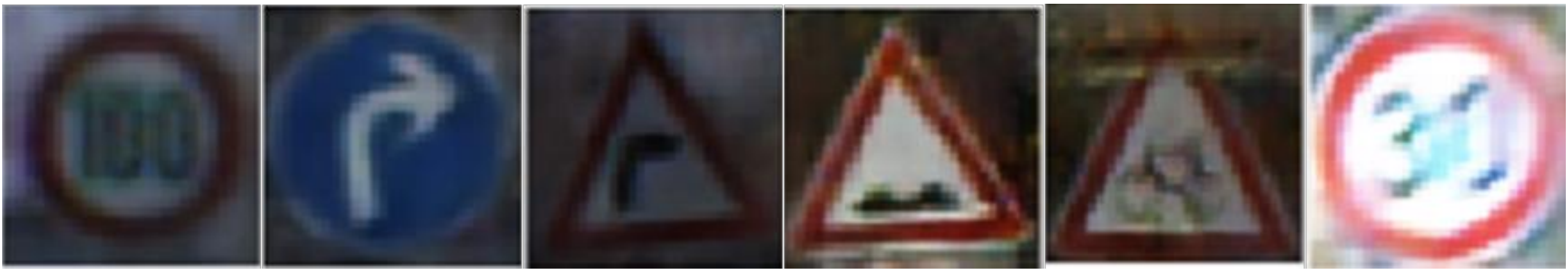}
\caption{Target's common part}
\label{fig:fig11f}
\end{subfigure}
\begin{subfigure}{0.47\textwidth}
\includegraphics[width=0.95\linewidth, height=0.95cm]{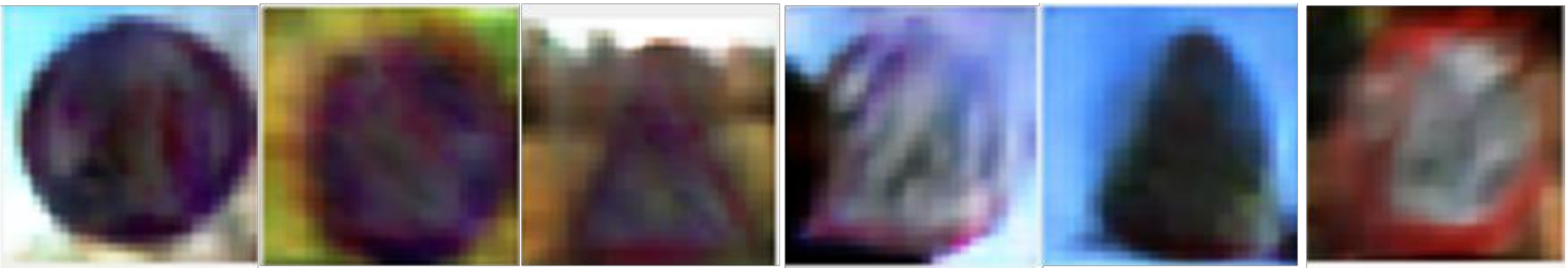}
\caption{Source's specific part}
\label{fig:fig11g}
\end{subfigure}
~~~
\begin{subfigure}{0.47\textwidth}
\includegraphics[width=0.95\linewidth, height=0.95cm]{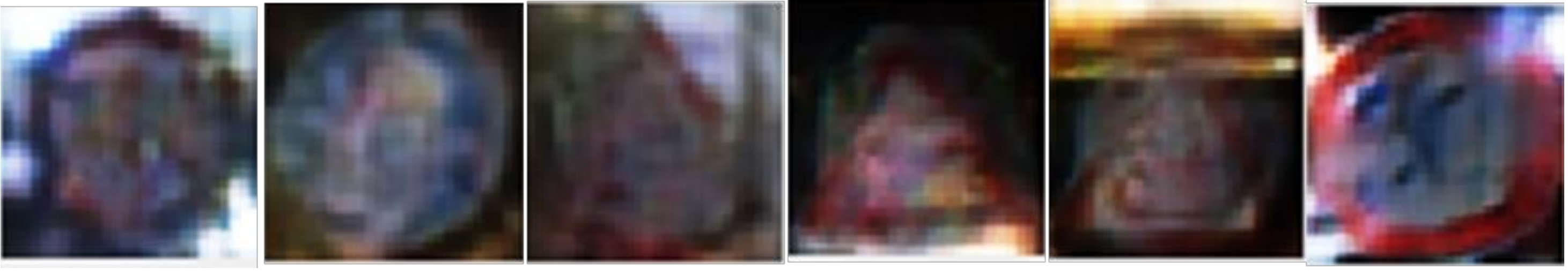}
\caption{Target's specific part}
\label{fig:fig11h}
\end{subfigure}
\caption{Syn-Signs $\rightarrow$ GTSRB}
\label{fig:fig11}
\end{center}
\end{figure}

In Fig.~\ref{fig:fig11} we show some examples for {\bf Syn-Signs $\rightarrow$ GTSRB} case. Here, we also visualize the common and specific parts in image domain by inhibiting the remaining part before inputting to the Generators. As shown in the Fig.~\ref{fig:fig11e} and \ref{fig:fig11f}, the common parts of each domain store the information of ``traffic signs''. Meanwhile, the specific parts encode the style information such as the background, shape, ... as shown in Fig.~\ref{fig:fig11g} and ~\ref{fig:fig11h}.

\end{document}